\documentclass[11pt]{article}

\makeatletter
\@ifundefined{papermode}{\def\papermode{preprint}}{}
\makeatother
\usepackage[\papermode]{acl}

\makeatletter
\newif\ifaclpreprint
\def\acl@preprint{preprint}
\ifx\papermode\acl@preprint \aclpreprinttrue \fi
\makeatother

\usepackage{times}
\usepackage{latexsym}
\usepackage[T1]{fontenc}
\usepackage[utf8]{inputenc}
\usepackage{microtype}
\usepackage{inconsolata}
\usepackage{graphicx}
\usepackage{amsmath,amssymb,bm}
\usepackage{booktabs}
\usepackage{multirow}
\usepackage{makecell}
\usepackage{tabularx}
\usepackage[table]{xcolor}
\usepackage{colortbl}
\usepackage{subcaption}
\usepackage{enumitem}
\usepackage{tikz}
\usetikzlibrary{positioning,calc,arrows.meta}
\usepackage{xspace}
\usepackage{placeins}
\usepackage[safe]{tipa}

\IfFileExists{latexml.sty}{\usepackage{latexml}}{\newif\iflatexml\latexmlfalse}

\definecolor{anuBlue}{HTML}{1A88FF}   
\definecolor{anuLink}{HTML}{0B6CD6}   
\definecolor{anuNavy}{HTML}{0B3F7E}
\definecolor{anuWash}{HTML}{ECF4FF}
\definecolor{anuChip}{HTML}{F3F4F6}
\definecolor{anuStack}{HTML}{C9D6E6}
\definecolor{anuInk}{HTML}{0A0A0A}
\definecolor{anuMuted}{HTML}{6E6E6E}
\definecolor{anuLine}{HTML}{BDBDBD}
\hypersetup{citecolor=anuLink, linkcolor=anuLink, urlcolor=anuLink}

\newcommand{\sys}{Anuvani-LID\xspace}
\newcommand{\suite}{SUITE-102\xspace}
\newcommand{\panel}{SUITE-42\xspace}
\newcommand{\pp}{\,pp\xspace}
\newcommand{\best}[1]{\textbf{#1}}
\newcommand{\ours}{\rowcolor{anuWash}}
\newcommand{\figpath}{figures/light}
\newcommand{\logopath}{figures/logos}
\newcommand{\hfmark}[1][1.2em]{\raisebox{-0.1em}{\includegraphics[height=#1]{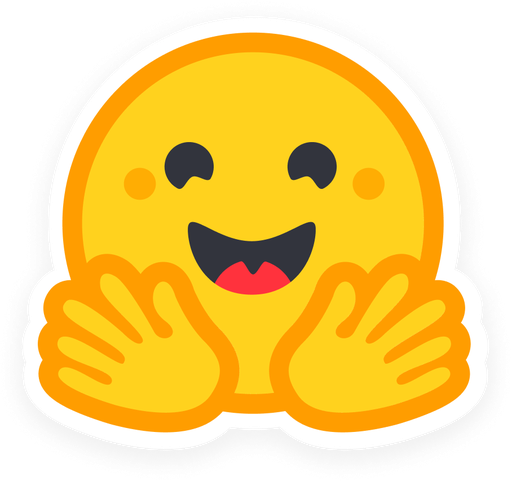}}}
\newcommand{\teasercaption}{\sys against the field on \panel (macro-F1, identical clips). The radar covers nine domains from 32 of the 37 core conditions, with its axis from 0.3; the leaderboard averages all 37; the bars show the four domains with the largest margin. The prompted APIs chose among 10 candidate languages, \sys among all 23.}
\iflatexml\else
\makeatletter
\AtBeginDocument{\g@addto@macro\@maketitle{%
  \vskip -0.37in
  \begin{minipage}{\textwidth}
    \centering
    \includegraphics[width=\textwidth]{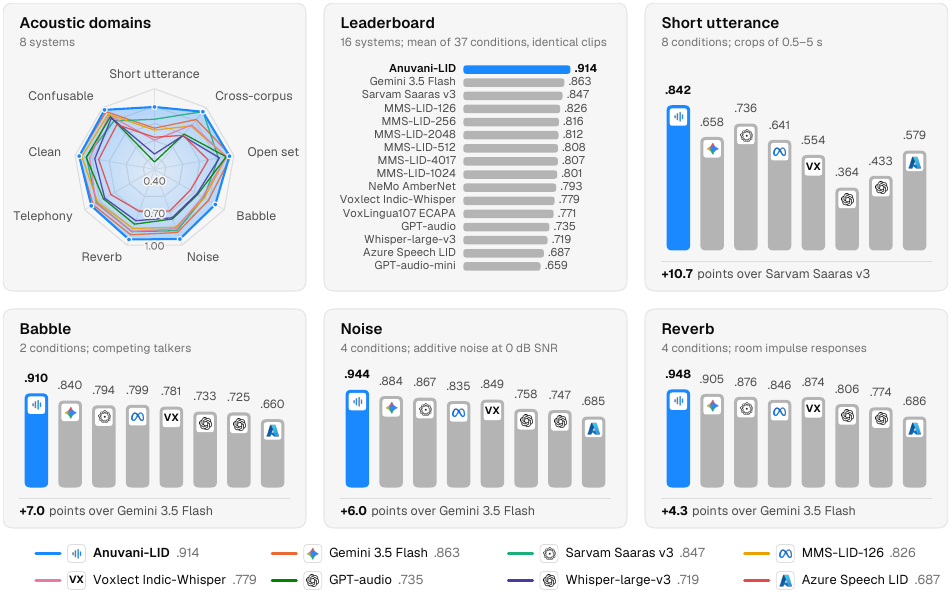}
    \captionof{figure}{\teasercaption}
    \label{fig:teaser}
  \end{minipage}
  \vskip 0.16in}}
\makeatother
\fi

\title{\sys: Robust and Calibrated Spoken Language Identification\\for Indian Languages}

\iflatexml
\author{Stephen Fernandes \\ Central Institute of Indian Languages \\ \texttt{stephenn.fernandes@gmail.com}
  \and Narayan Choudhary \\ Central Institute of Indian Languages \\ \texttt{n.choudhary@gov.in}}
\else
\author{\mdseries\itshape Stephen Fernandes, Narayan Choudhary \\[0.15em]
  \upshape Central Institute of Indian Languages \\[0.15em]
  {\normalsize\fontfamily{pcr}\selectfont stephenn.fernandes@gmail.com, n.choudhary@gov.in}}
\fi

\begin{document}
\maketitle

\begin{abstract}
  Every multilingual speech pipeline begins with a language decision, an error there propagates silently through recognition, translation and routing. In India it spans  22 scheduled languages across four language families, heard over short, noisy, narrowband and code-mixed audio. Foundational speech models increasingly present language identification (LID) as an inherent latent, generalized capability. Autoregressive decoders in speech models emit a special language ID token, and modern LLMs with audio capabilities can name the language when prompted. We test this assumption on identical clips and find that whisper and similar speech models has tokens for only 15 of the 23 languages, on one second inputs Gemini 3.5 Flash reaches 0.52 macro-F1 and GPT-audio 0.14. We present Anuvani LID, a dedicated 115M parameter LID conformer model trained on 54,643 hours of Indian speech spanning 22 Indian languages and Indian accented English. Along with designed and controlled studies on 102 condition benchmark with speaker and corpus held out tests. Across 37 conditions, it reaches 0.914 mean macro-F1, 5.2 points above Gemini 3.5 Flash and ahead of 14 other systems. It is 24 points ahead of Gemini at one second, scores 0.897 on all 23 languages where no other system exceeds 0.713, and is calibrated to an ECE of 0.008. Reliable Indian LID still requires a robust, dedicated and calibrated model.\ifaclpreprint\footnote{The released model is available at:\par\noindent\hfmark~\url{https://huggingface.co/CIIL/Anuvani-LID}.}\fi
\end{abstract}

\iflatexml
\begin{figure*}[t]
  \centering
  \includegraphics[width=\textwidth]{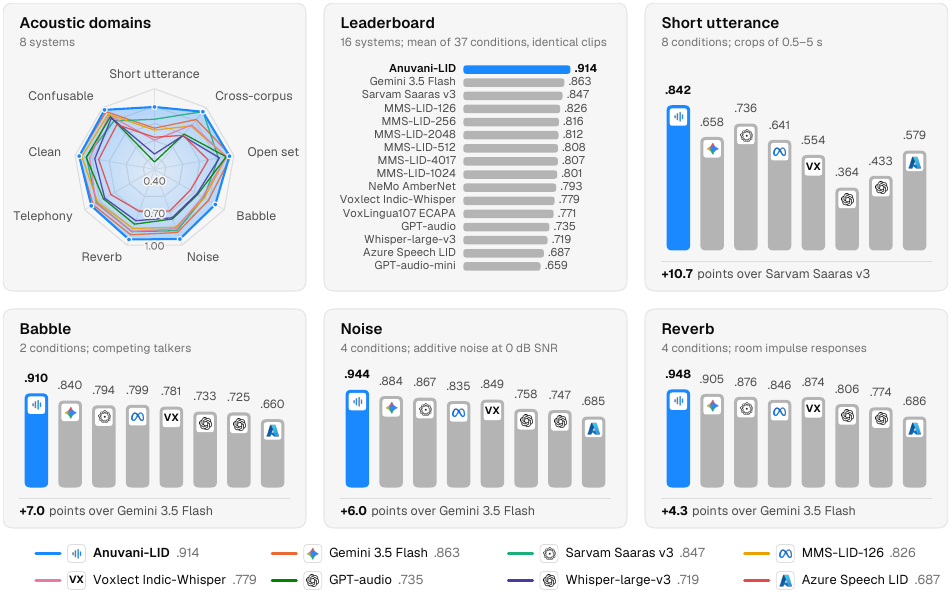}
  \caption{\teasercaption}
  \label{fig:teaser}
\end{figure*}
\fi

\section{Introduction}
\label{sec:intro}

Spoken language identification (LID) is the first decision in almost every multilingual speech pipeline: it chooses the recogniser, the translation direction and the language a helpline or assistant answers in \citep{li2013slr}. An error here propagates; a Maithili call sent to a Hindi recogniser degrades every later component. In India this decision is usually hard and complex. The Eighth Schedule lists 22 languages from four language families, the mother tongues of 96.7\% of the 1.21 billion people counted in 2011 \citep{orgi2018language}. Many are close relatives, our  training audio ranges from 13{,}039\,h (Indian accented English) to 369\,h (Kashmiri) in our pool, and deployed audio is short, noisy, narrowband and code-mixed (\S\ref{sec:ling}).

Speech foundational models represent LID task as a latent, in-context generalised capability. Whisper decoder decodes a language ID token before it transcribes \citep{radford2023whisper}, Meta's MMS models attaches a 4{,}017 language identification head \citep{pratap2024mms}, and LLMs with audio capabilities such as Gemini and GPT-audio name the language of a clip when asked \citep{google2026gemini35flash,openai2024gpt4o}. Dedicated LID models might therefore seem obsolete. For Indian languages, our measurements argue otherwise (\S\ref{sec:results}):
\begin{itemize}[leftmargin=*,itemsep=0pt,topsep=2pt]
  \item \textbf{Coverage is incomplete.} Whisper has tokens for only 15 of the 23 classes (22 scheduled languages plus English), and among open systems only Voxlect covers all 23.
  \item \textbf{Short inputs break generalist systems.} At one second audio length, Gemini 3.5 Flash reaches 0.52 macro-F1 and GPT-audio 0.14. At sub one second, most speech models fail.
  \item \textbf{Errors matter even at foundation-model scale.} Most of Whisper's supposed Welsh translation data turned out to be English audio misclassified by its language identifier \citep{radford2023whisper}. Even at the scale of Gemini 3.5 Flash, errors matter.
  \item \textbf{Deployment needs calibrated scores and small models.} Routing and rejection need calibrated confidence, which hard-label APIs often lack, and Whisper-large-v3 has 1.55B parameters \citep{openai2023whisperlargev3}, 13 times more than \sys.
\end{itemize}

A further obstacle is methodological. Clean, in-domain test sets saturate: in our pilot study every head reached about 98\% macro-averaged accuracy, within 1\pp of the others. Following work on saturation and shortcut learning \citep{geirhos2020shortcut,kiela2021dynabench}, we rebuilt the evaluation pipeline first and took every later decision on held-out, multi-axis conditions. We report the programme in the order it was carried out. Our contributions are:
\begin{itemize}[leftmargin=*,itemsep=0pt,topsep=2pt]
  \item \textbf{\suite}, a 102-condition, 318{,}861-utterance benchmark with speaker- and corpus-held-out tiers, and \textbf{\panel}, which scores 15 open and commercial systems on identical clips (\S\ref{sec:bench}).
  \item A \textbf{staged, controlled study} of fine-tuning a self-supervised Conformer for LID (\S\ref{sec:method}).
  \item \textbf{\sys}, which leads all 15 systems (Figure~\ref{fig:teaser}), most clearly on short, noisy and all-23-language conditions, and is well calibrated (\S\ref{sec:results}).
  \item A \textbf{linguistic error analysis} (\S\ref{sec:errors}) and a \textbf{layer analysis} showing that 12 of 17 blocks suffice (\S\ref{sec:layers}).
\end{itemize}

\section{The Linguistic Landscape of Indian Spoken LID}
\label{sec:ling}

\begin{table*}[t]
\centering
\scriptsize
\iflatexml\else 
\setlength{\tabcolsep}{1.7pt}
\renewcommand{\arraystretch}{0.95}
\fi
\begin{tabular}{@{}l l l r r r c @{\hspace{6pt}} l l l r r r c@{}}
\toprule
Language & Family: subgroup & Script(s) & L1 & Train & F1 & T & Language & Family: subgroup & Script(s) & L1 & Train & F1 & T \\
\midrule
Hindi & IA: Hindustani & Devanagari & 528.3 & 8{,}121 & .843 &  & Konkani$^{92}$ & IA: Southern & Devanagari, Latin & 2.3 & 375 & .918 &  \\
Urdu & IA: Hindustani & Arabic & 50.8 & 691 & .925 &  & Gujarati & IA: Guj.--Rajasthani & Gujarati & 55.5 & 1{,}500 & .948 &  \\
Punjabi & IA: Northwestern & Gurmukhi & 33.1 & 1{,}339 & .973 & T & Sanskrit & IA: Old Indo-Aryan & Devanagari & 0.02 & 828 & .962 &  \\
Dogri$^{04}$ & IA: W.\ Pahari & Devanagari & 2.6 & 478 & .977 & T & Tamil & Dravidian: South I & Tamil & 69.0 & 3{,}560 & .974 &  \\
Kashmiri & IA: Dardic & Arabic, Deva. & 6.8 & 369 & .953 &  & Malayalam & Dravidian: South I & Malayalam & 34.8 & 1{,}544 & .965 &  \\
Sindhi$^{67}$ & IA: Northwestern & Arabic, Deva. & 2.8 & 390 & .903 &  & Kannada & Dravidian: South I & Kannada & 43.7 & 3{,}287 & .975 &  \\
Nepali$^{92}$ & IA: N.\ (E.\ Pahari) & Devanagari & 2.9 & 1{,}150 & .961 &  & Telugu & Dravidian: South II & Telugu & 81.1 & 3{,}184 & .967 &  \\
Maithili$^{04}$ & IA: Bihari & Devanagari & 13.6 & 2{,}143 & .916 &  & Bodo$^{04}$ & Tib.-Burman: Bodo--Garo & Devanagari & 1.5 & 590 & .947 & T \\
Bengali & IA: Eastern & Bengali--Assamese & 97.2 & 3{,}517 & .972 &  & Manipuri$^{92}$ & Tib.-Burman: Meitei$^\ddagger$ & Meetei, Bengali & 1.8 & 539 & .933 & T \\
Assamese & IA: Eastern & Bengali--Assamese & 15.3 & 987 & .920 &  & Santali$^{04}$ & Austroas.: N.\ Munda & Ol Chiki & 7.4 & 394 & .935 &  \\
Odia & IA: Eastern & Odia & 37.5 & 1{,}827 & .899 &  & Indian English & IE: Germanic & Latin & 0.26$^\S$ & 13{,}039 & .901 &  \\
Marathi & IA: Southern & Devanagari & 83.0 & 4{,}792 & .941 &  &  &  &  &  &  &  &  \\
\bottomrule
\end{tabular}
\caption{The 23 classes. \emph{Family: subgroup} follows Glottolog 5.3 \citep{glottolog} and the Census \citep{orgi2018language} (IA: Indo-Aryan; IE: Indo-European; Arabic: Perso-Arabic; Deva.: Devanagari). \emph{Script(s)} follows CLDR and state official-language acts \citep{cldr48,unicode17}. \emph{L1}: first-language speakers in the 2011 Census (millions). \emph{Train}: hours in our training pool. \emph{F1}: \sys on the speaker-disjoint all-23 test. \emph{T}: documented lexical tone \citep{evans2018acoustics,das2019intonational,devi2024manipuri}. Superscripts give the year a language entered the Eighth Schedule if not an original member (1967, 1992, 2004; \citealp{goi2026constitution}). $^\ddagger$Subgrouping disputed \citep{thurgood2017subgroups}. $^\S$128.5M counting second- and third-language speakers \citep{orgi2011c17}; all English audio in training and test is Indian-accented.}
\label{tab:languages}
\end{table*}

\begin{figure}[t]
  \centering
  \includegraphics[width=\columnwidth]{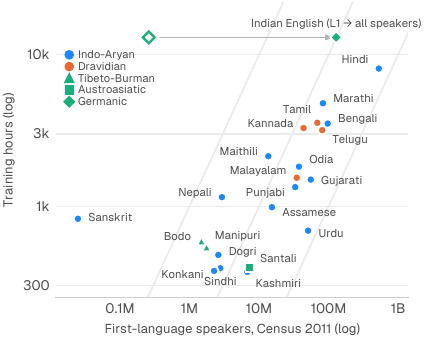}
  \caption{Speaker population versus training data. Speech resources do not track population: Urdu has more than 50M first-language speakers but less audio than Maithili or Nepali, and English is abundant because most of its 128.5M speakers use it as a second or third language. The diagonals mark 10, 100 and 1{,}000 training hours per million speakers.}
  \label{fig:languages}
\end{figure}

\paragraph{Twenty-two languages by design.} The Eighth Schedule for Indian languages grew from 14 to 22 languages through additions in 1967, 1992 and 2004 \citep{goi2026constitution,mha2017eighthschedule}. The 2011 Census rationalised 19{,}569 raw mother-tongue returns into 1{,}369 mother tongues and 121 languages \citep{orgi2018language}. Speaker numbers span four orders of magnitude, from Hindi (528M) to Sanskrit (24{,}821), and training data does not track them (Table~\ref{tab:languages}; Figure~\ref{fig:languages}). The labels are administrative: the Census ``Hindi'' groups 56 mother tongues, including Bhojpuri (50.6M), and Maithili was counted under Hindi until 2001 \citep{orgi2018language}, so class boundaries are partly conventional \citep{mha2017eighthschedule}.

\paragraph{Four families, one linguistic area.} Fifteen scheduled languages are Indo-Aryan, four Dravidian, two Tibeto-Burman and one Austroasiatic \citep{glottolog}. Genealogy understates how alike they sound. \citet{emeneau1956india} described India as a linguistic area in which languages of all families share retroflex consonants, verb-final clauses and echo words \citep{masica1976defining}, and retroflex systems are distributed geographically rather than genetically \citep{arsenault2017retroflexion}. Indian LID work reports heavily overlapping phone sets \citep{vuddagiri2018iiith,dey2022overview}, and on FLEURS the South Asian group has the lowest LID accuracy \citep{conneau2023fleurs}. Several classes are sisters or former dialects. Hindi and Urdu share colloquial Hindustani \citep{kachru2008hindi}, and Urdu$\to$Hindi is VoxLingua107's most common error \citep{valk2021voxlingua107}. Grierson treated Dogri as a dialect of Punjabi and Konkani as one of Marathi \citep{grierson1916lsi9a,grierson1905lsi7}, and these pairs share 93\% and 80--82\% of basic vocabulary \citep{kogan2016genealogical}. Script, which separates Hindi from Urdu (``one language, two scripts''; \citealp{king1994one}), is absent from speech. The remaining evidence is sparse in short clips: degree of retroflexion, laryngeal contrasts, vowel length, and lexical tone in Punjabi, Dogri, Bodo and Meitei.

\paragraph{Close relatives and dialect continua.} Several scheduled languages are sisters or former ``dialects'' of one another:
\begin{itemize}[leftmargin=*,itemsep=1pt,topsep=1pt]
  \item Hindi and Urdu are sister languoids under Hindustani and share a colloquial form \citep{glottolog,kachru2008hindi}. Urdu$\to$Hindi is the single most common error in VoxLingua107 \citep{valk2021voxlingua107}.
  \item Punjabi and Dogri share 93\% of basic vocabulary \citep{kogan2016genealogical}, and \citet{grierson1916lsi9a} described Dogri as a dialect of Punjabi.
  \item \citet{grierson1905lsi7} likewise treated Konkani as a Marathi dialect; the two share 80--82\% of basic vocabulary.
  \item Assamese, Bengali and Odia share 86--88\% of basic vocabulary and one Glottolog node, and they share numeral-classifier morphology with Maithili \citep{emeneau1956india}.
  \item Maithili speakers from the eastern dialect area commonly shift to Hindi when talking with speakers from other areas \citep{ciil_maithili_vari}.
\end{itemize}
Lexicostatistics measures vocabulary, not acoustics, but these are exactly the pairs where LID systems err (\S\ref{sec:errors}).

\paragraph{What the signal carries.} Script is absent from speech. Devanagari alone serves at least ten of the classes \citep{cldr48}. Hindi and Urdu, famously ``one language, two scripts'' \citep{king1994one}, are written in Devanagari and Perso-Arabic respectively. Labels must therefore come from corpus metadata rather than from transcripts. The discriminative evidence is phonetic, phonotactic and prosodic:
\begin{itemize}[leftmargin=*,itemsep=1pt,topsep=1pt]
  \item \textbf{Degree of retroflexion}: Tamil exceeds Hindi, and Nepali retroflexes are weaker still \citep{keane2004tamil,khatiwada2009nepali}.
  \item \textbf{Laryngeal series}: Hindi has a four-way stop series \citep{ohala1994hindi}, and Gujarati contrasts breathy and modal voice on vowels \citep{esposito2012contrastive}.
  \item \textbf{Vowel length}: contrastive in Dravidian languages but not in Bengali or Assamese \citep{keane2004tamil,khan2010bengali,mahanta2012assamese}.
  \item \textbf{Lexical tone}: documented in Punjabi and Dogri and in Bodo and Meitei \citep{evans2018acoustics,das2019intonational,devi2024manipuri}.
  \item \textbf{Rare segments}: Sindhi implosives and Santali checked finals \citep{maddieson2013glottalized,unicode17}.
\end{itemize}
Many of these cues are sparse, and a one-second clip may contain none of them. This is consistent with duration, rather than noise, being the dominant difficulty axis (\S\ref{sec:results}).

\paragraph{Indian English, not English.} The 23rd class is Indian accented English. Every English hour in our pool comes from Indian speakers, consisting of NPTEL lectures (10{,}441\,h), NLTM collections (1{,}252\,h), ULCA (916\,h) and Indic TIMIT (234\,h; \citealp{yarra2019indic}), and we evaluate on Indian-accented sets such as Svarah \citep{javed2023svarah}. \citet{kachru1986indianization} called Indian English ``an Indian language in its own right''; 128.5M Indians speak it, nearly all as a second or third language \citep{orgi2011c17}. Its accent is at once uniform and variable \citep{wiltshire2020uniformity,sailaja2009indian,pandey2015indian}. Pan-Indian features include apical, often retroflex /t d/, stops for /\textipa{T D}/, merged /v/--/w/ and unaspirated voiceless stops \citep{masica1972sound}, and a rhythm that differs from British English in duration, loudness, pitch and glottal stops \citep{fuchs2016speech}. Beneath them, the mother tongue shapes vowels, rhotics, diphthongs and pitch accents \citep{wiltshire2006influence,maxwell2010acoustic}. Svarah's 117 speakers from 65 districts cover 19 first languages, and Whisper's word error rate rises from 2.7\% on LibriSpeech to 7.2\% on it \citep{javed2023svarah}. For LID this cuts both ways. The English class must absorb all this variation, and each English utterance carries traces of the very languages the model must also recognise (\S\ref{sec:errors}).

\paragraph{Multilingual speakers.} In 2011, 26.0\% of Indians reported a second language and 7.1\% a third \citep{orgi2022atlas}. About 17\% of Hindi--English social-media posts are code-mixed \citep{bali2014borrowing}. In code-switched speech, more than half of the embedded segments last only 200--400\,ms \citep{barras2020vocapia}, so the matrix language is the natural utterance label.

\section{Related Work}
\label{sec:related}

\paragraph{Spoken LID.} LID moved from i-vectors \citep{li2013slr} to x-vectors \citep{snyder2018xvectors,snyder2018xvectorlid} and ECAPA-TDNN \citep{desplanques2020ecapa}, with VoxLingua107 \citep{valk2021voxlingua107} behind common open baselines \citep{speechbrain2021voxlingua107ecapa,jia2023ambernet}. Self-supervised (SSL) speech encoders now dominate the space with XLS-R and MMS type models as well as dedicate finetune's wav2vec~2.0-style models exists \citep{babu2022xlsr,pratap2024mms,baevski2020wav2vec2}, and \citet{bartley2023accidental} fine-tune an SSL Conformer. NIST LRE defines cluster-aware evaluation with $C_\text{avg}$ \citep{sadjadi2018nist,nist2022lreplan}.

\paragraph{Indian-language LID.} IIITH-ILSC \citep{vuddagiri2018iiith} introduced a 23-class benchmark. Later systems added CTC features \citep{mandava2019investigation} or fine-tuned XLSR-53 \citep{gupta2023enhancing,rajak2025deep}. Cross-corpus generalisation is the central difficulty for such models: EER rises from 11.8\% in-corpus to 43.3\% cross-corpus \citep{dey2021crosscorpora}, and acoustic augmentation and bias compensation recovers only part of it \citep{dey2023domain,dey2024towards}. Surveys and layer studies include \citet{dey2022overview} and \citet{dey2025probing}. On IndicSUPERB, 12-language LID accuracy falls from 98.2\% on seen to 90.8\% on unseen speakers \citep{javed2023indicsuperb}, and \citet{basu2026comparative} compare encoders for 42 Indic languages. We differ in scale (54.6k\,h), converging all 23 spoken language classes, held-out multi-axis evaluation, and comparison with commercial audio LLMs. We build upon the MahaDhwani encoder, which was pretrained w2v-BERT-style on 279k\,h of Indian audio \citep{bhogale2025mahadhwani,chung2021w2vbert}.

\section{\suite: A De-saturated Benchmark}
\label{sec:bench}


\paragraph{Design.} \suite probes LID along the multi-domain axes on which deployed systems fail (composition in Appendix Table~\ref{tab:suite}). Its headline tier is the IndicVoices test set \citep{javed2024indicvoices}: 300 spontaneous utterances per language for all 23 classes, from 1{,}308 speakers, none of whom is among the 34{,}034 IndicVoices speakers in our training data. Cross-corpus tiers use the evaluation splits of FLEURS, SPRING-INX, MUCS, Kathbath and Common Voice \citep{conneau2023fleurs,nithya2023springinx,diwan2021mucs,javed2023indicsuperb,ardila2020commonvoice}. The clip-level carve has an empty intersection with training, and transcript fingerprinting finds no FLEURS or MUCS overlap. Variety probes cover dialectal Hindi \citep{javed2024lahaja}, Hinglish \citep{singh2025hiacc}, Indian-accented English \citep{javed2023svarah}, studio speech \citep{baby2016indictts}, lectures \citep{ai4bharat2020nptel} and rural telephony \citep{bhanushali2022gramvaani}. Each source is expanded into random and VAD-anchored crops of 0.25--10\,s, MUSAN noise at 0 and 5\,dB \citep{snyder2015musan}, reverberation \citep{ko2017rir}, G.711 $\mu$-law telephony, babble, confusable-cluster subsets and open-set trials. The open-set trials use music, eight non-Indian languages and Chhattisgarhi, a relative of Hindi outside the label set. Curation is deterministic, and the suite was frozen before the final training run.

\paragraph{\panel.} Commercial APIs are too costly to run on the full suite, so we drew a balanced panel of 42 conditions with up to 150 utterances per language in each. The \emph{core} panel has 37 conditions (47{,}730 clips) over the ten languages every system supports: Hindi, Bengali, Tamil, Telugu, Marathi, Gujarati, Kannada, Malayalam, Punjabi and Urdu. The \emph{all-23} panel has five IndicVoices conditions (17{,}229 clips). Every system receives identical audio. Prompted audio LLMs and Azure receive the candidate list (10 or 23 languages), Sarvam Saaras runs in auto-detect mode, and open models predict over their native label sets. \sys always predicts over all 23 classes, so it can make errors that a 10-candidate system cannot. We also report it restricted to the same ten candidates (``matched'').

\paragraph{Metrics and statistics.} The primary metric is macro-F1 over the languages present in a condition. We also report top-1 accuracy, $C_\text{avg}$ \citep{nist2022lreplan}, EER, 15-bin ECE \citep{naeini2015obtaining,guo2017calibration}, and open-set AUROC from maximum softmax probability \citep{hendrycks2017baseline}. Intervals are 1{,}000-resample bootstraps \citep{bisani2004bootstrap} (300, stratified by condition, for paired differences), and paired items are tested with McNemar's test \citep{mcnemar1947note,gillick1989some} under Holm or Bonferroni correction \citep{holm1979simple}.

\section{Model and Staged Research Programme}
\label{sec:method}

\begin{figure*}[t]
  \centering
  \iflatexml 
  \includegraphics[width=\textwidth]{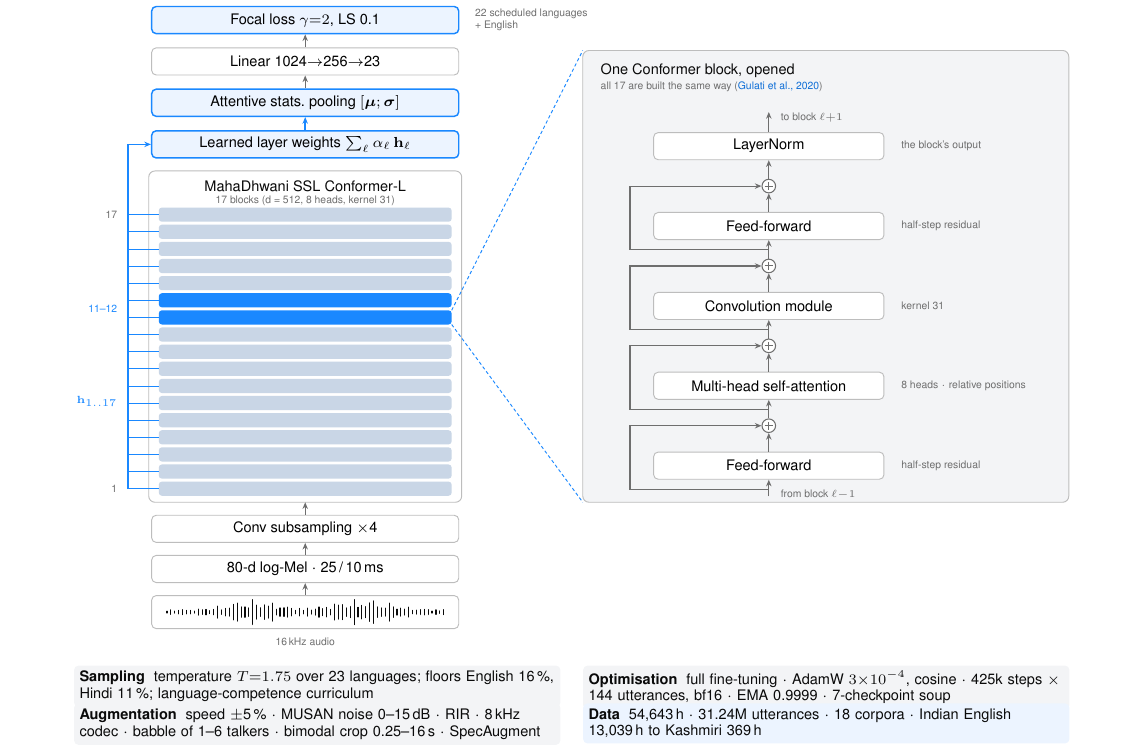}
  \else
  \resizebox{\textwidth}{!}{
\begin{tikzpicture}[
  font=\sffamily\scriptsize,
  box/.style={draw=anuLine, line width=0.45pt, rounded corners=2.2pt, fill=white, align=center,
              inner sep=2.6pt, minimum height=0.46cm, minimum width=5.2cm},
  hero/.style={draw=anuBlue, line width=0.7pt, rounded corners=2.2pt, fill=anuWash, align=center,
               inner sep=2.6pt, minimum height=0.46cm, minimum width=5.2cm},
  sub/.style={draw=anuLine, line width=0.45pt, rounded corners=2.2pt, fill=white, align=center,
              inner sep=2.2pt, minimum height=0.46cm, minimum width=3.9cm},
  chip/.style={draw=none, rounded corners=2pt, fill=anuChip, align=left, inner sep=2.6pt,
               font=\sffamily\scriptsize, text=anuInk},
  arr/.style={-{Stealth[length=3.2pt,width=2.6pt]}, line width=0.55pt, draw=anuMuted},
  harr/.style={-{Stealth[length=3.2pt,width=2.6pt]}, line width=0.7pt, draw=anuBlue},
  lbl/.style={font=\sffamily\tiny, text=anuMuted, align=center}
]
\def\cx{3.3}\def\W{5.2}
\node[box, minimum height=0.56cm] (wav) at (\cx,0.28) {};
\foreach \i in {0,...,71}{
  \pgfmathsetmacro{\env}{min(1, 0.22+0.78*exp(-((\i-22)/8.5)^2)+0.9*exp(-((\i-50)/9.5)^2))}
  \pgfmathsetmacro{\h}{0.215*\env*(0.30+0.70*abs(sin(\i*103)*cos(\i*41)))}
  \draw[anuInk, line width=0.55pt, line cap=round] ({\cx-2.34+\i*0.066},{0.28-\h}) -- ++(0,{2*\h});
}
\node[lbl, anchor=north] at (\cx,-0.03) {16\,kHz audio};
\node[box] (mel) at (\cx,1.01) {80-d log-Mel $\cdot$ 25\,/\,10\,ms};
\node[box] (sub) at (\cx,1.69) {Conv subsampling $\times$4};
\draw[arr] (wav) -- (mel); \draw[arr] (mel) -- (sub);
\def\sh{0.24}\def\sg{0.05}\def\fb{2.14}\def\bx{0.30}
\pgfmathsetmacro{\stop}{\fb+0.11+17*(\sh+\sg)-\sg}
\draw[anuLine, line width=0.45pt, rounded corners=3pt, fill=white]
  ({\cx-\W/2-0.05},\fb) rectangle ({\cx+\W/2+0.05},{\stop+0.62});
\foreach \i in {1,...,17}{
  \pgfmathsetmacro{\y}{\fb+0.11+(\i-1)*(\sh+\sg)}
  \def\fillc{anuStack}
  \ifnum\i>10 \ifnum\i<13 \def\fillc{anuBlue}\fi\fi
  \fill[\fillc, rounded corners=1.5pt] ({\cx-\W/2+0.12},\y) rectangle ({\cx+\W/2-0.12},{\y+\sh});
  \draw[anuBlue, line width=0.35pt] ({\cx-\W/2+0.12},{\y+\sh/2}) -- (\bx,{\y+\sh/2});
}
\node[font=\sffamily\scriptsize, text=anuInk, anchor=north] at (\cx,{\stop+0.58}) {MahaDhwani SSL Conformer-L};
\node[lbl, anchor=north] at (\cx,{\stop+0.33}) {17 blocks (d = 512, 8 heads, kernel 31)};
\node[font=\sffamily\tiny, text=anuMuted, anchor=east] at ({\bx-0.06},{\fb+0.11+\sh/2}) {1};
\node[font=\sffamily\tiny, text=anuBlue, anchor=east] at ({\bx-0.06},{\fb+0.11+10.5*(\sh+\sg)+\sh/2}) {11--12};
\node[font=\sffamily\tiny, text=anuMuted, anchor=east] at ({\bx-0.06},{\fb+0.11+16*(\sh+\sg)+\sh/2}) {17};
\node[font=\sffamily\tiny, text=anuBlue, anchor=east] at ({\bx-0.06},{\fb+0.11+5.5*(\sh+\sg)}) {$\mathbf{h}_{1..17}$};
\draw[arr] (sub.north) -- (\cx,\fb);
\pgfmathsetmacro{\ya}{\stop+0.62+0.22+0.23}
\node[hero] (agg) at (\cx,\ya) {Learned layer weights $\sum_\ell \alpha_\ell\, \mathbf{h}_\ell$};
\node[hero, above=0.22cm of agg] (asp) {Attentive stats.\ pooling $[\boldsymbol{\mu};\boldsymbol{\sigma}]$};
\node[box, above=0.22cm of asp] (fc) {Linear 1024$\to$256$\to$23};
\node[hero, above=0.22cm of fc] (loss) {Focal loss $\gamma{=}2$, LS 0.1};
\node[lbl, anchor=west, align=left] at ($(loss.east)+(0.14,0)$) {22 scheduled languages\\+ English};
\draw[anuBlue, line width=0.7pt] (\bx,{\fb+0.11+\sh/2}) -- (\bx,{\fb+0.11+16*(\sh+\sg)+\sh/2});
\draw[harr] (\bx,{\fb+0.11+16*(\sh+\sg)+\sh/2}) |- (agg.west);
\draw[harr] (agg) -- (asp); \draw[arr] (asp) -- (fc); \draw[arr] (fc) -- (loss);
\def\px{8.0}\def\pw{8.23}\def\pb{\fb}\pgfmathsetmacro{\pt}{\ya+3*0.46+2*0.22-0.23}
\draw[anuLine, line width=0.45pt, rounded corners=3pt, fill=anuChip] (\px,\pb) rectangle ({\px+\pw},\pt);
\node[font=\sffamily\scriptsize, text=anuInk, anchor=north west] at ({\px+0.18},{\pt-0.10}) {One Conformer block, opened};
\node[lbl, anchor=north west, align=left] at ({\px+0.18},{\pt-0.40}) {all 17 are built the same way \citep{gulati2020conformer}};
\def\sx{11.15}\def\rx{8.80}\pgfmathsetmacro{\yb}{\pb+0.62}\pgfmathsetmacro{\dy}{(\pt-1.62-\yb)/4}
\foreach \k/\nm/\dt in {0/Feed-forward/half-step residual, 1/Multi-head self-attention/8 heads $\cdot$ relative positions,
                        2/Convolution module/kernel 31, 3/Feed-forward/half-step residual, 4/LayerNorm/the block's output}{
  \node[sub] (m\k) at (\sx,{\yb+\k*\dy}) {\nm};
  \node[lbl, anchor=west] at ({\sx+2.12},{\yb+\k*\dy}) {\dt};
}
\foreach \k [evaluate=\k as \kn using int(\k+1)] in {0,...,3}{
  \node[circle, draw=anuMuted, line width=0.5pt, fill=white, inner sep=0pt, minimum size=0.23cm] (p\k)
    at (\sx,{\yb+\k*\dy+\dy/2}) {};
  \draw[anuMuted, line width=0.5pt] ($(p\k.center)+(-0.07,0)$) -- ++(0.14,0) ($(p\k.center)+(0,-0.07)$) -- ++(0,0.14);
  \draw[arr] (m\k.north) -- (p\k.south);
  \draw[arr] (p\k.north) -- (m\kn.south);
  \draw[arr] (\sx,{\yb+\k*\dy-0.40}) -| (\rx,{\yb+\k*\dy+\dy/2}) -- (p\k.west);   
}
\draw[arr] (\sx,{\pb+0.10}) -- (m0.south);
\node[lbl, anchor=west] at ({\sx+0.08},{\pb+0.13}) {from block $\ell{-}1$};
\draw[arr] (m4.north) -- ++(0,0.34);
\node[lbl, anchor=west] at ($(m4.north)+(0.08,0.24)$) {to block $\ell{+}1$};
\draw[anuBlue, dash pattern=on 1.6pt off 1.2pt, line width=0.45pt]
  ({\cx+\W/2-0.12},{\fb+0.11+10*(\sh+\sg)+\sh}) -- (\px,\pt)
  ({\cx+\W/2-0.12},{\fb+0.11+10*(\sh+\sg)}) -- (\px,\pb);
\node[chip, anchor=north west, text width=8.05cm] (c1) at (-0.62,-0.62) {\textbf{Sampling}\; temperature $T{=}1.75$ over 23 languages; floors English 16\,\%, Hindi 11\,\%; language-competence curriculum};
\node[chip, anchor=north west, text width=8.05cm] (c2) at (-0.62,-1.26) {\textbf{Augmentation}\; speed $\pm$5\,\% $\cdot$ MUSAN noise 0--15\,dB $\cdot$ RIR $\cdot$ 8\,kHz codec $\cdot$ babble of 1--6 talkers $\cdot$ bimodal crop 0.25--16\,s $\cdot$ SpecAugment};
\node[chip, anchor=north west, text width=8.05cm] (c3) at (8.0,-0.62) {\textbf{Optimisation}\; full fine-tuning $\cdot$ AdamW $3{\times}10^{-4}$, cosine $\cdot$ 425k steps $\times$ 144 utterances, bf16 $\cdot$ EMA 0.9999 $\cdot$ 7-checkpoint soup};
\node[chip, anchor=north west, text width=8.05cm, fill=anuWash] (c4) at (8.0,-1.26) {\textbf{Data}\; 54{,}643\,h $\cdot$ 31.24M utterances $\cdot$ 18 corpora $\cdot$ Indian English 13{,}039\,h to Kashmiri 369\,h};
\path ($(current bounding box.west)+(-1.147,0)$) ($(current bounding box.east)+(1.147,0)$);
\end{tikzpicture}}
  \fi
  \caption{\sys. All 17 block outputs of the self-supervised MahaDhwani Conformer are mixed with learned weights, pooled with attentive statistics and classified into 23 languages under focal loss. The whole network is fine-tuned. Blocks 11--12 (blue) carry the most linearly decodable language information (\S\ref{sec:layers}); the panel opens one block. Grey chips summarise the training recipe from Stages I--VI.}
  \label{fig:arch}
\end{figure*}

\subsection{Model}
\label{sec:model}
\sys fine-tunes the released MahaDhwani Conformer-L encoder \citep{bhogale2025mahadhwani,ai4bharat2024mahadhwanicard}, which was pretrained self-supervised on 279k\,h of large scale raw Indian audio (Figure~\ref{fig:arch}). The front-end uses 80-dimensional log-Mel features and 4$\times$ subsampling to 40\,ms frames. These feed 17 Conformer blocks \citep{gulati2020conformer} with $d{=}512$, 8 heads and convolution kernel 31. The head mixes all block outputs with learned weights, $\mathbf{z}_t=\sum_{\ell}\alpha_\ell \mathbf{h}_{\ell,t}$ with $\bm{\alpha}=\operatorname{softmax}(\mathbf{w})$ \citep{yang2021superb}. It then applies attentive statistics pooling \citep{okabe2018attentive}, a 256-dimensional embedding and a 23-way classifier (115.6M parameters in total). Training uses focal loss $-(1-p_y)^{\gamma}\log p_y$ \citep{lin2017focal} with $\gamma{=}2$ and label smoothing 0.1 \citep{szegedy2016rethinking}.

\subsection{A staged research programme}
The recipe is the outcome of six stages run in order, each starting from the previous stage's promoted recipe (Appendix Table~\ref{tab:stages}). Stages I--V ran at reduced scale (1{,}000--4{,}000\,h) so that many controlled arms were affordable; Stage VI trained once on all 54{,}643\,h, so that recipes here onwards could be validated on the full scale of training data.
Each stage answers one question with arms that change one axis at a time against the recipe it inherits, and architecture, sampling and augmentation are judged on held-out conditions (\S\ref{sec:bench}) rather than on in-domain validation accuracy, which saturates.

\begin{figure*}[t]
  \centering
  \includegraphics[width=\textwidth]{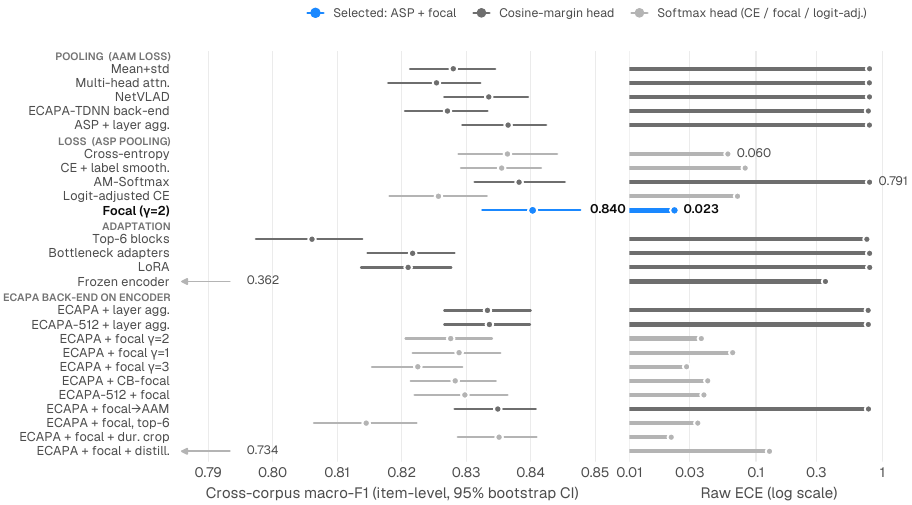}
  \caption{Stage~I (1{,}000\,h, 20 languages, 4 epochs, one run per arm): all 25 arms, the 11 ECAPA-TDNN back-end variants in the lowest group. Left: cross-corpus macro-F1 with 95\% bootstrap intervals. Right: raw ECE (log scale). Pooling choices differ by about 1\pp, but the loss decides calibration. No ECAPA arm improves on attentive pooling with focal loss (blue), despite 23M extra parameters.}
  \label{fig:arch_search}
\end{figure*}

\paragraph{Stage I: head, loss and adaptation search at 1{,}000\,h.} The first decision was the optimiser, because the heads under comparison differ in trainable size by two orders of magnitude, from LoRA adapters to an ECAPA-TDNN fine-tuned end to end. ULMFiT-style discriminative rates \citep{howard2018universal}, a larger step for the freshly initialised head than for the pre-trained encoder, are the usual prescription for this setting. We tested them against a single global rate on three heads that span the range, attentive pooling, NetVLAD and ECAPA-TDNN, for 1{,}500 steps each. A head rate ten times the encoder's ($10^{-3}$ against $10^{-4}$) gave the large NetVLAD head nothing (0.980 against 0.982 macro-averaged validation accuracy) and destabilised the angular-margin attentive head, so a single global rate of $10^{-4}$, at which the pre-trained encoder and the new head co-adapt stably, became the default for every fully fine-tuned arm.

We trained 25 arms on a balanced 1{,}000\,h subset of 20 languages (Kashmiri, Santali and Sindhi had too little audio) and scored them on the 54-condition predecessor of \suite (Figure~\ref{fig:arch_search}). Every arm shared one recipe apart from the axis under test: AdamW \citep{loshchilov2019decoupled} with $\beta{=}(0.9,0.98)$ and weight decay $10^{-3}$, 10\% linear warm-up and cosine decay to $10^{-6}$, 144 utterances per step, temperature sampling at $T{=}2$, bf16 and four epochs.
Three findings shaped the model:
\begin{itemize}[leftmargin=*,itemsep=1pt,topsep=1pt]
  \item \textbf{Pooling matters little.} Under an additive angular margin (AAM) loss, statistics \citep{snyder2018xvectors}, multi-head \citep{india2019self}, NetVLAD \citep{arandjelovic2016netvlad}, ECAPA-TDNN \citep{desplanques2020ecapa} and attentive pooling over aggregated layers all land within 0.825--0.837 cross-corpus macro-F1, with overlapping intervals.
  \item \textbf{The loss decides calibration.} Cosine-margin heads \citep{deng2019arcface,wang2018additive} emit bounded, rescaled cosines, and fine-tuned cosine-margin heads have an ECE of 0.75--0.79; softmax heads reach 0.02--0.08 (0.13 under distillation). With attentive pooling, focal loss has the highest macro-F1 of all 25 arms (0.840) and an ECE of 0.023, 2.6$\times$ below cross-entropy, in line with \citet{mukhoti2020calibrating}; layer aggregation, however, classifies 1.1\% more items correctly on top-1, so we kept both.
  \item \textbf{The encoder must be fine-tuned.} A frozen encoder reaches only 0.362, and top-6 fine-tuning (0.806), adapters (0.822; \citealp{houlsby2019parameter}) and LoRA (0.821; \citealp{hu2022lora}) all trail full fine-tuning. These partial regimes ran at the larger rates prescribed for small trainable budgets, $10^{-3}$ for the frozen encoder's head, for bottleneck adapters of width 64 and for LoRA of rank 8 ($\alpha{=}16$), and $3{\times}10^{-4}$ for the top six blocks, and the frozen encoder trails full fine-tuning by 48 points.
\end{itemize}
ECAPA-TDNN \citep{desplanques2020ecapa}, explored in 12 arms, was set aside: its deep head drives the arccos-based ArcFace to divergence, it over-corrects under focal loss and under self-distillation \citep{hinton2015distilling}, and no ECAPA arm passes attentive pooling with focal loss on cross-corpus macro-F1 (0.835 against 0.840) despite 23M extra parameters. Fusing arms promises up to 11.1 points on short utterances under an oracle, but majority vote recovers almost none of it, so we kept a single network, the simplest arm at the top of the cross-corpus ranking: attentive statistics pooling, focal loss and full fine-tuning.

\begin{figure}[t]
  \centering
  \includegraphics[width=\columnwidth]{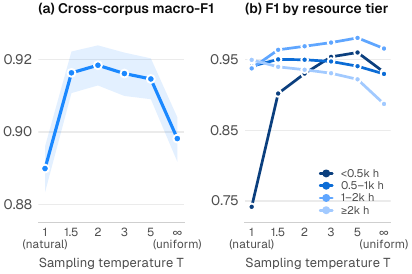}
  \caption{Stage II (4{,}000\,h, production skew, 1 epoch). (a) Cross-corpus macro-F1 against sampling temperature, with a 95\% band. (b) In-domain F1 by resource tier, with tiers set by hours in the full pool; the in-domain test has no Kashmiri, Santali or Sindhi, so its lowest tier is Dogri and Konkani. Raising $T$ trades head for tail.}
  \label{fig:sampling}
\end{figure}

\paragraph{Stage II: rebalancing under production skew.} On 4{,}000\,h that preserve the large scale language distribution, in terms of multilingual language ratio, orthographic and phonetic diversity, this ensures that our proxy training corpus exactly represents the large scale production training corpus and provides us results that translate at scale during larger production training runs. we sampled languages with $p_l\propto n_l^{1/T}$ \citep{arivazhagan2019massively,conneau2020unsupervised}. Cross-corpus macro-F1 follows an inverted U, peaking at $T{=}2$ (0.919, against 0.890 at $T{=}1$ and 0.898 uniform; Figure~\ref{fig:sampling}).
The replica applies one fraction, 11.18\%, to every language $\times$ corpus $\times$ duration stratum of the production corpus as it stood before its final expansion to 54{,}643\,h, so it keeps all three marginals at once: no language's share moves by more than 0.005 points, and the corpus mix by at most 0.002 in $L_1$ distance. Sixteen arms change one axis each against an attentive-pooling, focal-loss spine trained for one epoch.
Raising $T$ trades head for tail. Between $T{=}1$ and $T{=}2$ the two languages under 500\,h that the in-domain test covers, Dogri and Konkani, rise from 0.742 to 0.930 mean F1 (Dogri alone from 0.50 to 0.89), while the eight languages above 2{,}000\,h in the full pool slip only from 0.949 to 0.935. Temperature is therefore a policy choice between head and tail, not a quantity to maximise.
In-domain macro-F1 traces the same inverted U, from 0.923 at $T{=}1$ to 0.946 at $T{=}2$ and 0.947 at $T{=}3$, falling back to 0.922 under uniform sampling. Seen- and unseen-speaker scores stay within half a point of each other up to $T{=}2$ and separate by about 1.5 points only at $T{=}5$ and under uniform sampling, which repeat the tail's few speakers too often; this is an independent reason to stop at $T{\approx}2$.

\begin{itemize}[leftmargin=*,itemsep=1pt,topsep=1pt]
  \item \textbf{Temperature sampling gains with training length.} The $T{=}2$ recipe trained for two epochs instead of one improves every axis: in-domain macro-F1 rises from 0.946 to 0.958, pooled cross-corpus macro-F1 from 0.919 to 0.931, confusable pairs from 0.885 to 0.909, degraded audio from 0.832 to 0.852 and open-set AUROC from 0.829 to 0.840. The second epoch lifts every resource tier, the two sub-500\,h languages from 0.930 to 0.970 and the eight largest from 0.935 to 0.943, so a longer schedule recovers more than half of the head accuracy that a single epoch of up-sampling had traded away.
  \item \textbf{Focal loss complements temperature sampling.} The two act on different quantities. Temperature sampling sets how often each language is seen; focal loss sets how much each example counts once seen, scaling its loss by $(1-p_y)^{\gamma}$ so that confidently classified examples fade and hard ones dominate the gradient. The sampler corrects class frequency and the loss corrects example difficulty, so neither undoes the other. Stage~I measured what focal loss adds on its own: on the same attentive head it raises cross-corpus macro-F1 by 0.86 points over the angular-margin loss while cutting ECE from 0.788 to 0.023, it nearly triples cross-corpus recall of Dogri, the hardest tail language (0.075 to 0.20), and against cross-entropy it is 2.6 times better calibrated at slightly higher accuracy (0.840 against 0.836). Within focal loss, $\gamma{=}1$ matches $\gamma{=}2$, and $\gamma{=}3$ gains 1.0 point cross-corpus and 3.6 points at 1\,s but raises ECE to 0.078, so $\gamma{=}2$ stays as the calibrated setting.
\end{itemize}
Rebalancing must happen once, in the sampler or in the loss. As the sole rebalancer, logit adjustment \citep{menon2021longtail} at $T{=}1$ matches temperature sampling on the four plain cross-corpus sets (0.925 against 0.925 at $T{=}3$) and is the best-calibrated arm of the stage (ECE 0.011 against 0.046 for plain focal loss at $T{=}1$). On top of $T{=}2$ it counts the prior twice: in-domain macro-F1 falls 3.5 points and the two largest classes collapse, English from 0.906 to 0.699 F1 and Hindi from 0.713 to 0.344. A two-stage schedule, one epoch of uniform sampling followed by one of natural sampling, lifts in-domain macro-F1 to 0.964 against 0.958 for two single-stage epochs, but weakens open-set separation (AUROC 0.798 against 0.840), so the single-stage sampler stayed. At equal budget, learned layer aggregation raises in-domain macro-F1 from 0.958 to 0.967 and improves every acoustic family (telephony +2.8\pp), so we adopted it; it also lowers $100\cdot C_\text{avg}$ from 2.23 to 1.72.

\begin{figure}[t]
  \centering
  \includegraphics[width=\columnwidth]{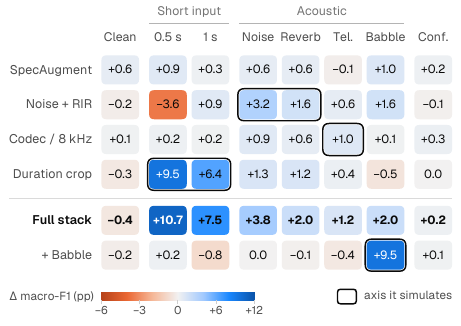}
  \caption{Stage III: change in macro-F1 (points) relative to a no-augmentation baseline, on the 44 conditions common to all arms (4{,}000\,h, 2 epochs). The babble row is measured against the full stack without babble (65k steps). Each perturbation mainly moves the axis it simulates. The full stack combines their gains at a 0.4-point cost on clean speech.}
  \label{fig:augmentation}
\end{figure}

\paragraph{Stage III: augmentation.} Each perturbation mainly improves the axis it simulates (Figure~\ref{fig:augmentation}):
\begin{itemize}[leftmargin=*,itemsep=1pt,topsep=1pt]
  \item random duration cropping: +9.5\pp at 0.5\,s and +6.4\pp at 1\,s;
  \item MUSAN noise with room impulse responses: +3.2\pp in noise and +1.6\pp in reverberation and babble (alone it costs 3.6\pp at 0.5\,s);
  \item $\mu$-law/8\,kHz codec simulation: +1.0\pp on telephony;
  \item SpecAugment: small, uniform gains;
  \item multi-talker babble, 1--6 overlapping talkers at 5--18\,dB SNR, on top of the other four: +11.6\pp on the seven babble conditions (0.804 to 0.920 macro-F1), with no cost on clean conditions (+0.3\pp).
\end{itemize}
Combined, the other four gain 10.7\pp at 0.5\,s, 7.5 at 1\,s, 3.8 in noise, 2.0 in reverberation and babble, and 1.2 on telephony, for a 0.4\pp cost on clean speech. The final recipe crops half of the training segments. Half of those crops are drawn log-uniformly between 0.25 and 3\,s and the rest uniformly between 3 and 16\,s, which concentrates training mass on the short durations that interactive systems receive.

Eight operations make up the final recipe (Table~\ref{tab:augmentation}). Six act on the waveform as each clip leaves the data loader, always in the same order: speed perturbation \citep{ko2015audio}, additive noise from MUSAN \citep{snyder2015musan}, reverberation with simulated room impulse responses \citep{ko2017rir}, the duration crop, babble and the codec. Noise therefore enters before the room response and is reverberated with the speech, babble is mixed in at the cropped length, and the telephone channel acts last, on the whole mixture, as it does in a call. The other two are the frequency and time masks of SpecAugment \citep{park2019specaugment}, applied to the log-Mel features of every training example. Each waveform transform is selected independently for every clip, with a fixed probability that depends on neither the language nor the corpus, so no transform can become a cue for a class, and its parameters are drawn afresh each time. The six selections give 64 combinations. Over the 61.2M training examples of the final run, 102.5k\,h of source audio, 93.5\% pass through at least one waveform transform and 6.5\% reach the masks unaltered. A clip receives 2.1 waveform transforms on average, and 70.3\% receive two or more. Speed perturbation picks one of five rates from 0.95 to 1.05 and leaves the clip as it is at the identity rate, so it changes 40\% of clips. Half of the clips are selected for cropping, but a crop from the long branch, uniform between 3\,s and the clip's own length, cannot shorten a clip of 3\,s or less, so 44.9\% are shortened; a quarter of all examples, 15.3M, are cut to between 0.25 and 3\,s. Noise comes from 930 MUSAN recordings, and reverberation from 60{,}000 impulse responses simulated in 600 rooms, 200 each small, medium and large. Babble sums one to six talkers drawn from 10{,}426 recordings: the 426 speech recordings of MUSAN, as in x-vector training \citep{snyder2018xvectors}, and 10{,}000 training clips spread over all 23 languages. The codec band-limits to 4\,kHz by resampling through 8\,kHz and applies 8-bit G.711 $\mu$-law companding \citep{itu_g711}. The Stage~III arms used the same probabilities, with a single uniform crop of 0.5--16\,s, and babble only in its own arm.

\begin{table}[t]
\centering
\small
\setlength{\tabcolsep}{3pt}
\begin{tabular}{@{}l l r r r r@{}}
\toprule
 & & & \multicolumn{3}{c}{Altered} \\
\cmidrule(l){4-6}
Operation & Setting & $p$ & \% & M & k\,h \\
\midrule
Speed & 0.95--1.05 & 0.5 & 40.0 & 24.5 & 41.0 \\
Noise & 0--15\,dB & 0.5 & 50.0 & 30.6 & 51.2 \\
Reverberation & 60k RIRs & 0.3 & 30.0 & 18.4 & 30.7 \\
Duration crop & 0.25--16\,s & 0.5 & 44.9 & 27.5 & 49.6 \\
Babble & 5--18\,dB & 0.2 & 20.0 & 12.2 & 20.5 \\
Codec & 8\,kHz $\mu$-law & 0.3 & 30.0 & 18.4 & 30.7 \\
\midrule
Frequency masks & $2\times{\le}20$ bins & 1 & 100 & 61.2 & 102.5 \\
Time masks & $2\times{\le}5$\% & 1 & 100 & 61.2 & 102.5 \\
\midrule
Any waveform op. & & & 93.5 & 57.2 & 96.3 \\
None & & & 6.5 & 4.0 & 6.2 \\
\bottomrule
\end{tabular}
\caption{Augmentation in the final run, in the order of application. $p$ is the probability that a clip is selected for the operation. \emph{Altered} gives the share of training examples the operation changes, their number in millions and the thousands of hours of source audio they carry, as expected totals over the 61.2M examples of the 425k-step run, computed from the language mix the run's sampler drew and the clip durations of its training index.}
\label{tab:augmentation}
\end{table}

\paragraph{Stage IV: curriculum learning.} Stages I--III fix what the model sees; Stage IV asks whether the order in which it sees it adds more \citep{bengio2009curriculum}. We expected the largest return where the static recipe left most error: on the confusable Indo-Aryan cluster, the weakest clean axis of Stage II (0.909 macro-F1 on confusable pairs), and on sub-second input. Following the decomposition of a curriculum into a difficulty measure and a training scheduler \citep{wang2022survey}, we screened four definitions of difficulty under one competence schedule \citep{platanios2019competence}, $p(e)=\min\bigl(1,\sqrt{e\,(1-p_0^2)/(N-1)+p_0^2}\bigr)$ with $p_0{=}0.2$, which moves an axis from its easy end to its hard end over $N$ epochs. Each arm changed one axis of the Stage III recipe (4{,}000\,h, three epochs) and had to lift its target without regressing any resource tier or language:
\begin{itemize}[leftmargin=*,itemsep=1pt,topsep=1pt]
  \item \textbf{Duration.} Training crops shrink from 6.6 to 2.95 to 1.0\,s, replacing the static random crop. This inverts SortaGrad \citep{amodei2016deep}: a short clip is easy for recognition but carries less evidence for LID.
  \item \textbf{Resource.} The sampling temperature rises from 1.2 to 2.0, so the head dominates early and the tail gains weight late, after the annealed language sampling of mmBERT \citep{marone2025mmbert}; a reverse schedule, from 1.8 to 1.0, controls for direction.
  \item \textbf{Confusability.} The 15 Indo-Aryan languages enter at 0.36 of their temperature-sampled weight, rising to 0.78 and then 1.0, so the Dravidian, Tibeto-Burman and Austroasiatic languages and English are seen at full weight first and the cluster in which most confusions arise (\S\ref{sec:errors}) is brought in gradually.
  \item \textbf{Self-paced.} Within each batch only the lowest-loss 68\% of examples enter the loss, rising to all of them \citep{kumar2010self}.
\end{itemize}
Ordering by duration gives the largest short-input gain of any curriculum. Against the unscheduled recipe trained for the same three epochs, it lifts mean macro-F1 over five corpora from 0.42--0.43 to 0.52 at 0.5\,s and from 0.17--0.18 to 0.26 at 0.25\,s. Because ending training on one-second crops also shortens what the model hears of full-length speech, the final recipe takes the same short-crop exposure from the bimodal crop of Stage~III, which keeps long segments in every batch. The model-driven gate did not add to the unscheduled recipe.

\begin{figure*}[t]
  \centering
  \includegraphics[width=\textwidth]{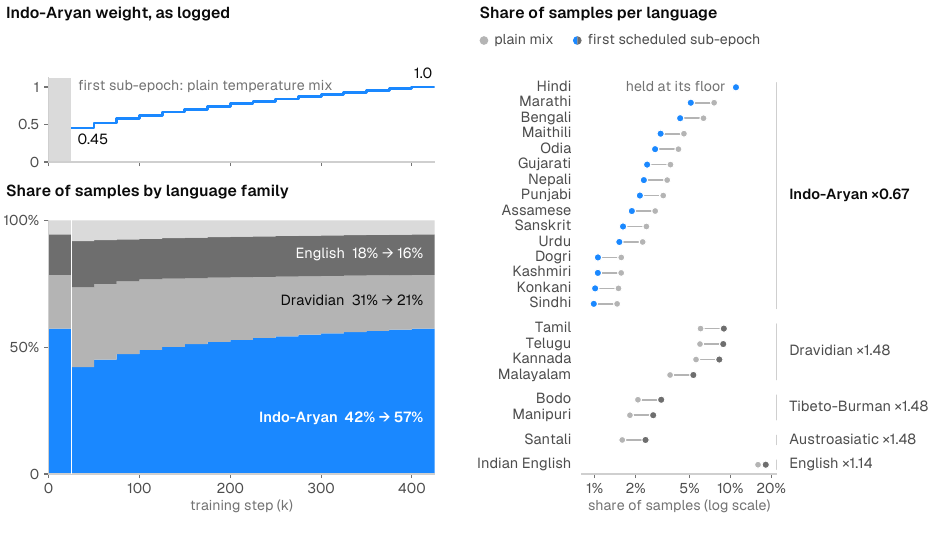}
  \caption{Stage IV: the confusability curriculum as the final run applied it (425k steps, 17 sub-epochs of 25k). Top left: the weight on the 15 Indo-Aryan languages, as logged; the first sub-epoch draws the plain temperature mix. Bottom left: the share of samples by language family. Right: the share of samples per language in the plain mix and in the first scheduled sub-epoch, with each family's scale factor; Hindi is held at its 11\% floor. Shares come from the run's own sampler on its training index and reproduce every share it logged.}
  \label{fig:curriculum}
\end{figure*}

The final recipe schedules the language mix. It keeps the confusability curriculum, which acts on the cluster that carries most errors, composes with temperature sampling by changing which languages fill a batch as training proceeds rather than how often each is drawn relative to its size, and costs nothing at inference. At full scale its competence ramp spans the 17 sub-epochs of 25k steps that make up the 425k-step budget: after a first sub-epoch on the plain temperature mix, the weight of the Indo-Aryan cluster rises from 0.45 to 1.0 (Figure~\ref{fig:curriculum}), on top of $T{=}1.75$ and the language floors of Stage~V. The schedule is applied in the streaming shard reader exactly as in the reference sampler (the language distributions agree to $10^{-13}$), and the final run logs it at every sub-epoch. We chose this curriculum on the evidence of the ablations. Confusable pairs were the weakest clean axis of Stage~II (0.909 macro-F1), and the Indo-Aryan cluster carries most of the remaining errors: 65--66\% of those of the unscheduled three-epoch recipe on the all-23 test. Each of the other three curricula either acts elsewhere or would move a tuned setting. The bimodal crop of Stage~III already supplies the short-crop exposure of the duration curriculum while keeping long segments in every batch, the model-driven gate did not add to the unscheduled recipe, and the resource curriculum keeps the sampling temperature away from $T{=}2$, where the inverted U of Stage~II peaks, for most of training. The confusability curriculum alone targets the dominant error while keeping temperature sampling and the language floors as tuned, and it returns to the tuned mix by the last sub-epoch.

\paragraph{Stage V: scale-aware optimisation.} Short probes on the 4{,}000\,h proxy set the optimiser. This proxy was redrawn from the final 54{,}643\,h pool with the Stage II construction, so its language ratios match the full pool. Learning rates of $3{\times}10^{-5}$ to $10^{-3}$ gave development accuracies of 0.771--0.876; we took the knee at $3{\times}10^{-4}$, since the last step added only 0.5\pp. Short horizons flatter large rates, so we checked the choice at length: at $3{\times}10^{-4}$ development accuracy reached 0.941 after 25k steps, a level the earlier $10^{-4}$ recipe needed about 46k steps to reach. Warm-ups of 5--15\% and temperatures of 1.5--2.5 were flat within 1.2\pp, so we chose $T{=}1.75$ with floors keeping English at 16\% and Hindi at 11\% of samples. $T{=}1.75$ is the value at which the full augmentation stack, babble included, was validated, and because the proxy keeps the pool's language ratios, the same temperature yields the same sampling mix at full scale; only the number of repetitions grows. The floors protect the two largest classes. Temperature sampling alone would give English 12.3\% and Hindi 10.6\% of draws, and under focal loss, which discounts easy examples, that leaves the head too little exposure; the Hindi floor also guards the Hindi--Urdu boundary. We sized the budget by effective passes: over 425k steps of 144 utterances, the five lowest-resource languages are seen four to six times each and the high-resource languages 1.3 to 2.2 times. With the tail replayed several times, label smoothing of 0.1 \citep{szegedy2016rethinking} guards against memorisation. To feed this budget, a sequential shard reader streams the corpus from a columnar store: it cuts read amplification from 38$\times$ to 0.97$\times$ and delivers 449 clips/s per data worker against a training demand of about 180, so the GPUs, not the disks, set the pace.

\begin{figure}[t]
  \centering
  \includegraphics[width=\columnwidth]{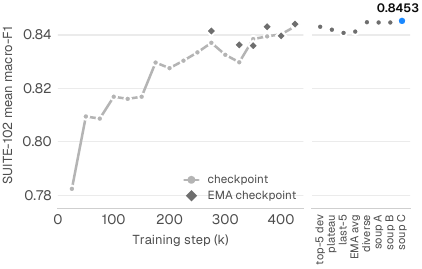}
  \caption{Stage VI (54{,}643\,h). Left: mean macro-F1 on \suite of the checkpoints saved every 25k steps and of their exponential moving averages (EMA). Right: uniform weight averages (``soups'') of checkpoint subsets. The 7-checkpoint soup ranks first of 31 candidates.}
  \label{fig:trajectory}
\end{figure}

\paragraph{Stage VI: full-scale training and weight averaging.} The final run trains on all 54{,}643\,h (31.24M utterances) for 425k steps with AdamW \citep{loshchilov2019decoupled}, cosine decay, the confusability curriculum of Stage~IV, bf16 and an EMA of the weights with decay 0.9999, on six RTX A6000 GPUs (Appendix~\ref{app:hparams}). It took 69.6\,h, and development accuracy rose to 0.951 at the final checkpoint. \suite macro-F1 climbs from 0.782 at 25k steps to 0.844 for the final EMA weights (Figure~\ref{fig:trajectory}), but the condition families do not peak together. Clean speech peaks at 75k steps, telephony at 175k, confusable pairs and open-set rejection at 200k, and babble, noise, reverberation and short utterances only at 425k. A single late checkpoint therefore keeps the optima reached at the end of training and gives up the earlier ones. Telephony, for instance, falls from 0.932 at its peak to 0.898 at the final step, and the EMA, with a horizon of about 10k steps, averages only the end of the run. We therefore finalise the model as recent large-scale training recipes do, by averaging weights along the training trajectory \citep{izmailov2018averaging,wortsman2022model}. Llama~3 averages checkpoints from its annealing phase \citep{grattafiori2024llama}, and OLMo~2 averages independently annealed runs \citep{olmo2024olmo2}. We archived a checkpoint every 25k steps beside its EMA shadow and scored 31 candidates on all 102 conditions: the 17 checkpoints, six late EMA shadows, five averages over contiguous or diverse subsets, and three soups whose members were fixed from the per-family profiles before any soup was scored. Every average is uniform over parameters and normalisation statistics, accumulated in float64. The best candidate, a soup of seven checkpoints at 100k, 175k, 275k, 325k, 350k, 400k and 425k steps, scores 0.8453 and is the released \sys. Against the final checkpoint, it lifts telephony from 0.898 to 0.923 and babble, noise and open-set rejection by 0.6, 0.4 and 0.4 points. It also lowers raw ECE on the all-23 test from 0.077 to 0.054, while no condition family falls by more than 0.5 points. Early members bring back the channel robustness that late training erodes. Because the average is taken in weight space, unlike the output-space fusion of Stage~I, it costs nothing at inference.

\section{Comparison with Open and Commercial Systems}
\label{sec:results}

\begin{table}[t]
\centering
\footnotesize
\iflatexml\else\setlength{\tabcolsep}{1.5pt}\fi 
\begin{tabular}{@{}l c c c r c@{}}
\toprule
 & & Core-37 & All-23 & \multicolumn{1}{c}{Gap} & \\
System & Lang. & F1 & F1 & \multicolumn{1}{c}{{\scriptsize[95\% CI]}} & W/L \\
\midrule
\ours \textbf{\sys} & 23 & \best{.914} & \best{.897} & \multicolumn{1}{c}{---} & --- \\
\ours \quad matched & 10 & .923 & .897 & & \\
\midrule
Gemini 3.5 Flash & 23 & .863 & .710 & 5.2 {\scriptsize[4.8, 5.5]} & 21/9 \\
Sarvam Saaras & 23 & .847 & .346 & 6.8 {\scriptsize[6.5, 7.1]} & 26/5 \\
GPT-audio & 23 & .735 & .384 & 18.0 {\scriptsize[17.5, 18.4]} & 34/2 \\
Azure LID & 11 & .687 & --- & 22.7 {\scriptsize[22.1, 23.3]} & 36/1 \\
GPT-audio-mini & 23 & .659 & .285 & 25.5 {\scriptsize[25.1, 26.0]} & 34/2 \\
\midrule
MMS-LID-126 & 15 & .826 & .686$^*$ & 8.8 {\scriptsize[8.4, 9.3]} & 31/1 \\
MMS-LID-256 & 19 & .816 & .533$^*$ & 9.8 {\scriptsize[9.4, 10.3]} & 28/1 \\
MMS-LID-2048 & 21 & .812 & .478$^*$ & 10.2 {\scriptsize[9.8, 10.7]} & 28/1 \\
MMS-LID-4017 & 21 & .807 & .467$^*$ & 10.7 {\scriptsize[10.3, 11.2]} & 30/1 \\
AmberNet & 15 & .793 & .633$^*$ & 12.2 {\scriptsize[11.7, 12.6]} & 34/1 \\
Voxlect Indic & 23 & .779 & .713 & 13.5 {\scriptsize[13.1, 14.0]} & 35/0 \\
VL107 ECAPA & 15 & .771 & .637$^*$ & 14.3 {\scriptsize[13.8, 14.8]} & 34/0 \\
Whisper-v3 & 15 & .719 & .482$^*$ & 19.5 {\scriptsize[19.1, 20.0]} & 35/1 \\
\bottomrule
\end{tabular}
\caption{\panel results on identical clips. Lang.\ counts how many of the 23 classes a system can output. \emph{Gap} is the paired core macro-F1 difference, \sys minus the system (points), with a 95\% interval from a bootstrap stratified by condition. W/L counts core conditions on which \sys is significantly better or worse (McNemar, Bonferroni-corrected over 37). $^*$Over the covered languages only, which favours partial-coverage systems. Top-1 accuracy and MMS-LID-512/1024 are in Appendix~\ref{app:panel}.}
\label{tab:panel}
\end{table}

\paragraph{Overall.} On the 37 core conditions of \panel, \sys reaches 0.914 mean macro-F1 while choosing among all 23 classes (Table~\ref{tab:panel}). That is 5.2 points above Gemini 3.5 Flash \citep{google2026gemini35flash} and 6.8 above Sarvam Saaras v3 \citep{sarvam2026saaras}; restricted to the APIs' ten candidates it reaches 0.923 (+6.0 over Gemini, 95\% CI [5.7, 6.4]). The best open model, MMS-LID-126 \citep{pratap2024mms}, trails by 8.8 points. Every paired difference excludes zero (McNemar $p<10^{-100}$ over 47{,}730 clips). Per condition, \sys is significantly better than Gemini on 21 of 37 and worse on 9, and better than every open model on at least 28.

\begin{figure*}[t]
  \centering
  \includegraphics[width=\textwidth]{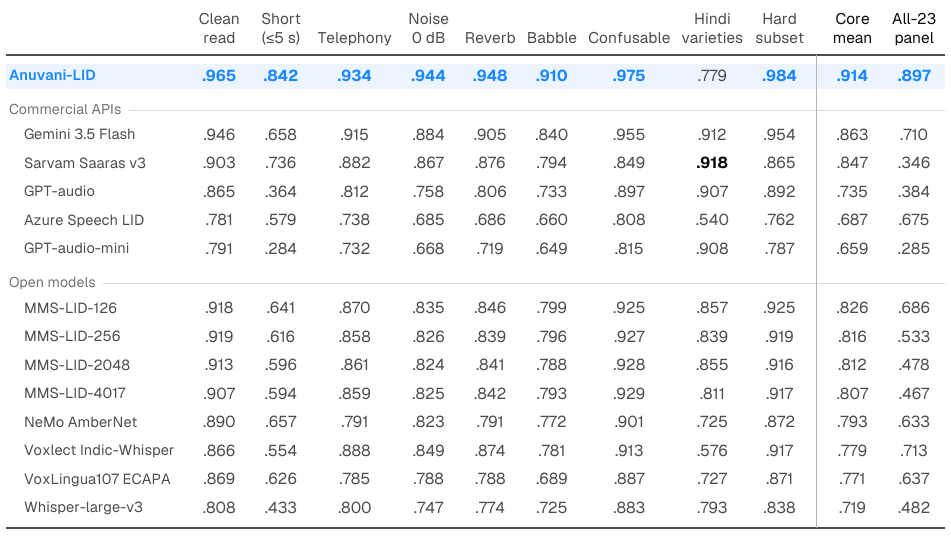}
  \caption{Mean macro-F1 by condition domain on \panel (identical clips; \sys scored over 23 classes). Bold marks the best system in each column, in blue where it is \sys. Hindi varieties are rural telephony (GramVaani), dialectal Hindi (Lahaja) and Hinglish (HiACC); see \S\ref{sec:errors}.}
  \label{fig:domains}
\end{figure*}

\paragraph{Where the margin comes from.} \sys has the best mean in eight of nine domains (Figure~\ref{fig:domains}). Its margin over Gemini is largest on short utterances (0.842 against 0.658), babble (+7.0), noise at 0\,dB (+6.0) and reverberation (+4.3), and smallest on clean read speech (+1.9) and telephony (+1.8). It holds cross-corpus on SPRING-INX (0.953 against 0.865). Gemini is significantly better on six long read-speech conditions, the five full-length FLEURS slices and MUCS, by 1.2--5.7 points (Appendix~\ref{app:panel}); read sentences such as FLEURS's Wikipedia text \citep{conneau2023fleurs} offer lexical content an audio LLM can exploit. It is also better on two Hindi-variety probes (\S\ref{sec:errors}).

\paragraph{Short utterances.} Duration separates systems most. At one second \sys reaches 0.764 macro-F1 on FLEURS and 0.854 in-domain, 24 points above Gemini in both cases (0.520, 0.611). At 0.5\,s it reaches 0.545, against 0.439 for Sarvam and 0.231 for Gemini. Duration cropping (Stage III) plausibly explains the gap; some competitors never saw short audio, and Voxlect, for example, was trained on 3--15\,s clips \citep{feng2025voxlectindic}.

\begin{figure*}[t]
  \centering
  \includegraphics[width=\textwidth]{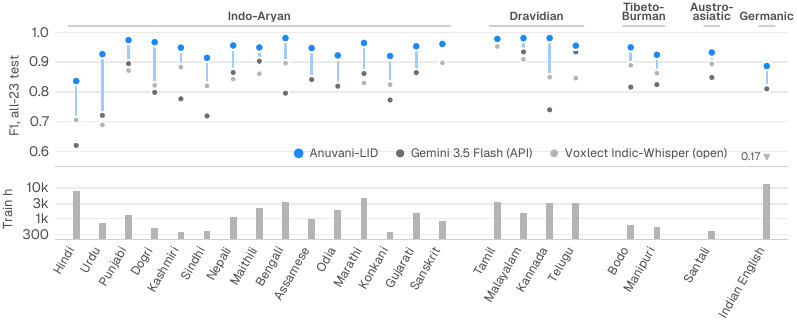}
  \caption{Per-language F1 on the all-23 panel (IndicVoices, speaker-disjoint, 150 clips per language, identical for all systems), grouped by family, with training hours below. Blue connectors show the gain over the better rival.}
  \label{fig:per_language}
\end{figure*}

\paragraph{All 23 languages.} Only five other systems can in principle output all 23 classes. On the all-23 panel \sys averages 0.897 macro-F1, against 0.713 for Voxlect Indic-Whisper, 0.710 for Gemini and 0.384 for GPT-audio. Its F1 exceeds 0.90 on 21 languages, including Bodo, Dogri, Santali, Kashmiri and Sindhi with under 600\,h each, and is lowest for Hindi (0.84) and Indian English (0.89) (Figure~\ref{fig:per_language}). Gains over the better of Gemini and Voxlect are largest for Urdu (+20.5), Dogri (+14.5), Hindi and Kannada (+13.1); only on Tamil is a rival marginally ahead (0.984 against 0.977).

\paragraph{The full suite against open models.} On the 73 \suite conditions where an open model covers at least eight core languages, \sys scores highest on 71, with a mean margin of 12.8 points over the best open model; the exceptions are FLEURS telephony (MMS-LID-2048, +2.4) and IndicTTS (AmberNet, +0.4).

\begin{figure}[t]
  \centering
  \includegraphics[width=\columnwidth]{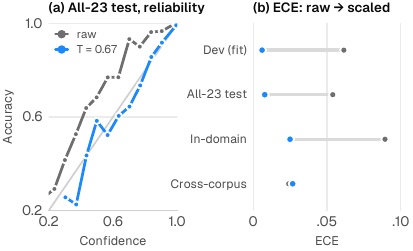}
  \caption{Calibration. (a) Reliability on the all-23 test before (grey) and after (blue) temperature scaling fitted on the development set. (b) ECE on four sets.}
  \label{fig:calibration}
\end{figure}

\paragraph{Calibration, open set and selection.} One temperature ($T{=}0.674$) fitted on the development set \citep{guo2017calibration} lowers ECE on the held-out all-23 test from 0.054 to 0.008 (Figure~\ref{fig:calibration}); there, $100\cdot C_\text{avg}$ is 3.16 and EER 1.79\%. Maximum-softmax rejection reaches AUROC 0.89--0.91 for music and non-Indian languages but 0.84 for Chhattisgarhi, a close relative of Hindi. The unselected final EMA checkpoint scores 0.912 core macro-F1 and 0.889 all-23, so selecting the soup on \suite changes no ranking.

\section{Linguistic Error Analysis}
\label{sec:errors}

\begin{figure}[t]
  \centering
  \includegraphics[width=0.9\columnwidth]{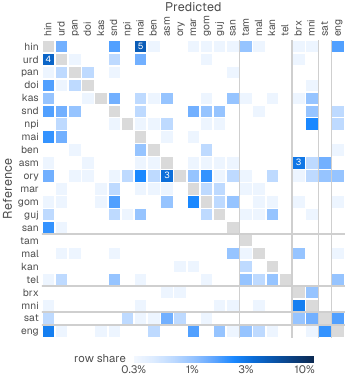}
  \caption{Off-diagonal confusions of \sys on the speaker-disjoint all-23 test (6{,}900 utterances), as row shares on a log scale, grouped by family as in Table~\ref{tab:languages}. Cells of 3\% or more are annotated.}
  \label{fig:confusion}
\end{figure}

The residual errors of \sys are linguistically structured (Figure~\ref{fig:confusion}). On the all-23 test, 61\% of errors stay within the reference language's family, and 73\% originate in Indo-Aryan languages.

\paragraph{Sisters within a subgroup.} The largest confusions link the close relatives of \S\ref{sec:ling}: Hindi$\to$Maithili (4.7\%) and Maithili$\to$Hindi (2.0\%), a pair the Census merged until 2001; Urdu$\to$Hindi (4.0\%) within Hindustani; Dogri to Hindi and Punjabi; Konkani to Marathi (2.3\%); and Odia to Assamese (3.3\%) and Maithili (2.3\%) within the Magadhan cluster. Hindi, whose Census label covers 56 mother tongues, is the hardest class (F1 0.84).

\paragraph{Neighbours across family lines.} Other errors cross family but not geographic boundaries. Assamese$\to$Bodo (3.0\%), Manipuri$\to$Bodo (2.7\%), Nepali$\to$Manipuri (2.3\%) and Santali$\to$Assamese involve languages in long contact in north-eastern India; Assamese, for example, has many Bodo-related words \citep{mahanta2012assamese}. Such areal similarity \citep{emeneau1956india,thomason2000linguistic} is invisible in a family tree, though shared recording regions may also contribute.

\paragraph{L1 transfer in Indian English.} English is confused mainly with Hindi (2.7\%), Santali (2.0\%) and Marathi (1.7\%). This matches the L1-conditioned phonology of Indian English \citep{masica1972sound,wiltshire2006influence}. The same effect is stronger on Svarah, whose 117 speakers cover 19 first languages \citep{javed2023svarah}: 40\% of its accented-English utterances are attributed to a scheduled language, spread over many classes.

\paragraph{Hindi varieties.} The one \panel domain where other systems lead is Hindi varieties, and here too \sys errs along linguistic lines: 19\% of dialectal Hindi (Lahaja) is labelled Urdu, and Hinglish (HiACC) goes 28\% to Urdu, which shares Hindi's spoken grammar, and 23\% to English, which supplies its embedded words. These single-language probes reward any prior towards the probed class, and the prompted APIs could not answer ``English''. Under the matched protocol, \sys's HiACC macro-F1 rises from 0.551 to 0.729, and on the balanced confusable conditions it leads Gemini (0.975 against 0.955). Colloquial Hindi and Urdu share one form \citep{kachru2008hindi}, so the choice between them is largely sociolinguistic, and code-mixed speech calls for frame-level language diarization \citep{baghel2023displace,dey2022overview}.

\section{Where Is Language Identity Encoded?}
\label{sec:layers}

\begin{figure*}[t]
  \centering
  \includegraphics[width=\textwidth]{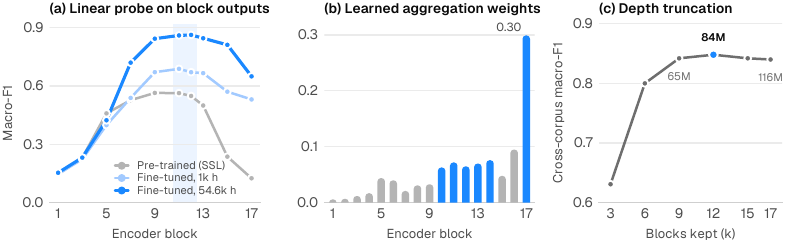}
  \caption{Where language identity lives. (a) Macro-F1 of linear probes on pooled block outputs of the pre-trained encoder, a 1{,}000\,h fine-tune and the 54.6k\,h model; the band marks blocks 11--12. (b) Softmax-normalised layer weights learned by the 54.6k\,h model. (c) Cross-corpus macro-F1 after truncating the encoder to its first $k$ blocks (Stage-I baseline recipe, 1{,}000\,h, 2 epochs; parameter counts annotated; two seeds for $k \geq 12$).}
  \label{fig:layers_full}
\end{figure*}

We probe pooled block outputs \citep{belinkov2022probing}, inspect the learned aggregation weights and truncate the encoder (Figure~\ref{fig:layers_full}), and we knock out single blocks. The full-scale subject is the first 54.6k\,h checkpoint of the recipe.

\paragraph{Findings.} \emph{Language lives mid-to-upper stack.} In the pre-trained encoder, probe macro-F1 peaks at about 0.56 in blocks 9--11 and collapses to 0.13 at block 17, whose representations serve the pre-training objective \citep{pasad2021layerwise}. Fine-tuning moves the peak to block 11 (0.69) at 1{,}000\,h and block 12 (0.86) at 54.6k\,h, higher than the lower-layer optimum reported for an SSL Conformer by \citet{bartley2023accidental}. \emph{Aggregation buys robustness.} At full scale block 17 receives 30\% of the learned weight and blocks 10--16 most of the rest; no single-block knockout then costs more than 0.08 macro-F1, whereas bypassing block 7 costs a last-block model 0.34. \emph{Channel is not the cue.} Corpus identity stays linearly decodable at the top (0.90; majority 0.25), yet erasing its rank-3 subspace with LEACE \citep{belrose2023leace} changes LID macro-F1 by only +0.1 points, so the decision does not rest on the channel cues behind earlier cross-corpus failures \citep{dey2021crosscorpora}. \emph{Twelve blocks suffice.} Truncation saturates at $k{=}12$ (0.848 with 84M parameters against 0.840 with all 17 blocks and 116M; 0.844 against 0.835 at four epochs with two seeds).

\section{Conclusion}
\label{sec:conclusion}

Language identification is not yet a solved by-product of scale. For India's 22 scheduled languages and Indian English, the generalist systems we tested lack classes, collapse on short inputs or return uncalibrated labels. A dedicated 116M-parameter model, designed on held-out, multi-axis conditions, does better on each count. Its residual errors follow the genealogy and contact history of India's languages, making code-switching, dialect continua and unseen relatives the next frontier.

\section*{Limitations}
\paragraph{Label granularity.} \sys assigns one label per utterance. For code-switched speech such as Hinglish this is ill-posed. Frame-level language diarization \citep{baghel2023displace} is the appropriate formulation, and our HiACC and Lahaja probes show that single-label scoring penalises acoustic models on mixed and dialectal Hindi.

\paragraph{Coverage.} The label set covers the 22 scheduled languages and Indian English. India's many other languages (Bhili, Gondi, Tulu, Bhojpuri and others) are outside it. Our open-set results show that rejecting close relatives of scheduled languages, such as Chhattisgarhi, is harder than rejecting unrelated languages.

\paragraph{Evaluation.} \suite includes one licensed in-domain comparator (LDC-IL; \citealp{choudhary2021ldcil}). All held-out generalisation tiers use public corpora and can be rebuilt from our scripts. Kathbath shares 4 of 456 speakers with training and is therefore reported only as an in-domain comparator. Commercial APIs return hard labels, so we compare them on macro-F1 and accuracy but not on calibration or $C_\text{avg}$, and their behaviour may change between versions. We report the API versions and dates of our runs in Appendix~\ref{app:systems}.

\paragraph{Variance and scale.} Stages I--V used single runs per arm. We measured run-to-run variance and interpret effects accordingly, but small within-stage differences remain provisional. Stage VI was trained once. Its checkpoint trajectory and the unselected-checkpoint check indicate that the conclusions do not hinge on selection, but we did not repeat full-scale training with other seeds. Sub-second LID remains information-limited. At 0.25\,s \sys reaches 0.24--0.35 macro-F1 in seven of the eight 0.25\,s conditions, and no open model exceeds 0.18 where it can be scored.

\section*{Ethics Statement}
All training and evaluation data come from published corpora under their licences. The one licensed comparator is used for evaluation only. We use no private user audio: the production-mix condition is a held-out sample of the public training corpora in their deployment proportions. Language identification can be misused to profile speakers by language, region or community. It is also a prerequisite for providing services in citizens' own languages, and it is where current systems are least reliable for speakers of low-resource scheduled languages. We therefore report per-language results for all 23 classes, and we do not recommend using LID output as a proxy for ethnicity, religion or nationality. Hindi and Urdu, in particular, share a spoken register, so an utterance's label should not be read as a statement about its speaker's identity. Where speaker gender is recorded, about one third of the training speech is female. We did not evaluate gender-sliced performance in this paper and flag it as necessary future work. Training the final model took about three days on six GPUs.

\begingroup 
\microtypesetup{expansion=false}
\bibliography{references}
\endgroup

\clearpage
\appendix
\makeatletter
\setlength{\@fptop}{0pt}\setlength{\@fpsep}{12pt plus 4pt}\setlength{\@fpbot}{0pt plus 1fil}
\setlength{\@dblfptop}{0pt}\setlength{\@dblfpsep}{12pt plus 4pt}\setlength{\@dblfpbot}{0pt plus 1fil}
\makeatother
\renewcommand{\topfraction}{0.9}\renewcommand{\dbltopfraction}{0.9}\renewcommand{\textfraction}{0.07}
\renewcommand{\floatpagefraction}{0.6}\renewcommand{\dblfloatpagefraction}{0.6}
\section{Benchmark composition}
\label{app:suite}

Table~\ref{tab:suite} breaks \suite down by source tier and condition family. Speaker-held-out tiers share no speaker with training. Split-held-out tiers use the evaluation split of a corpus whose training split we used, and their clip-level intersection with training is empty. Corpus-held-out probes come from corpora absent from training. Clip-held-out tiers are unseen clips of training corpora and serve as in-domain comparators. The \suite score is the unweighted mean macro-F1 over all 102 conditions.

\begin{table*}[t]
\centering
\small
\setlength{\tabcolsep}{4pt}
\begin{tabular}[t]{@{}l l r r r@{}}
\toprule
Source tier & Held out by & Lang. & Cond. & Utts \\
\midrule
IndicVoices, all-23 test & speaker (0/1{,}308) & 23 & 12 & 78.6k \\
IndicVoices, cross-corpus & speaker & 21 & 13 & 40.2k \\
FLEURS & split & 13 & 16 & 41.4k \\
Kathbath & split & 12 & 16 & 41.8k \\
Common Voice & split & 8 & 12 & 16.1k \\
MUCS & split & 6 & 12 & 14.0k \\
SPRING-INX & split & 10 & 6 & 12.0k \\
Production mix & clip & 22 & 2 & 41.5k \\
In-domain (LDC-IL) & clip & 20 & 7 & 27.9k \\
Variety probes$^\dagger$ & corpus / split & 1--9 & 6 & 3.4k \\
\midrule
Total & & 23 & 102 & 317.0k \\
\bottomrule
\end{tabular}\hspace{3em}%
\begin{tabular}[t]{@{}l r@{}}
\toprule
Condition family & Cond. \\
\midrule
Duration crops, 0.25--10\,s & 34 \\
VAD-anchored crops & 6 \\
Plain & 17 \\
Noise, 0 and 5\,dB SNR & 14 \\
Reverberation & 7 \\
8\,kHz $\mu$-law telephony & 7 \\
Babble & 7 \\
Confusable clusters & 7 \\
Open-set & 3 \\
\midrule
Total & 102 \\
\bottomrule
\end{tabular}
\caption{Composition of \suite. Utterance counts are in-set clips; the three open-set conditions add 1{,}900 out-of-set negatives, for 318{,}861 utterances in total. $^\dagger$HiACC (Hinglish code-switching), Lahaja (dialectal Hindi), Svarah (Indian-accented English), IndicTTS, NPTEL lectures and GramVaani (rural telephony).}
\label{tab:suite}
\end{table*}

\section{Research programme and Stage~VI settings}
\label{app:hparams}

Table~\ref{tab:stages} summarises the six stages of \S\ref{sec:method}. Figure~\ref{fig:augmentation} gives the complete Stage~III results; the complete Stage~I results are in Figure~\ref{fig:arch_search}. Table~\ref{tab:hparams} lists the settings of the full-scale run, and Figure~\ref{fig:trajectory} shows its checkpoint trajectory. The released model is a uniform average of seven checkpoints from this trajectory, which were saved every 25k steps: members 3, 6, 10, 12, 13, 15 and 16, i.e.\ steps 100k, 175k, 275k, 325k, 350k, 400k and 425k. Inference uses a single temperature, $T{=}0.674$, fitted on the 3{,}450-utterance development set.

\begin{table*}[t]
\centering
\small
\setlength{\tabcolsep}{5pt}
\begin{tabular}{@{}l l r l@{}}
\toprule
Stage & Data (classes) & Arms & Decision carried forward \\
\midrule
0 Benchmark & --- & --- & held-out, multi-axis suite \\
I Head, loss, adaptation & 1k\,h balanced (20) & 25 & attentive statistics pooling, focal loss, full fine-tuning \\
II Rebalancing & 4k\,h production skew (23) & 16 & $T{\approx}2$; a single rebalancer; learned layer mix \\
III Augmentation & 4k\,h production skew (23) & 8 & duration crops, noise, RIR, codec, babble \\
IV Curriculum & 4k\,h production skew (23) & 5 & confusability ramp of the 15 Indo-Aryan languages \\
V Optimisation & 4k\,h proxy (23) & 15 & peak LR $3{\times}10^{-4}$; $T{=}1.75$ with floors \\
VI Full scale & 54.6k\,h (23) & 31 cand. & EMA; 7-checkpoint soup \\
\bottomrule
\end{tabular}
\caption{The research programme in chronological order.}
\label{tab:stages}
\end{table*}

\begin{table*}[t]
\centering
\small
\setlength{\tabcolsep}{4pt}
\begin{tabular}{@{}>{\raggedright\arraybackslash}p{0.2\textwidth} p{0.74\textwidth}@{}}
\toprule
Setting & Value \\
\midrule
Front-end & 80-d log-Mel, 25\,ms window, 10\,ms hop; 4$\times$ convolutional subsampling \\
Encoder & MahaDhwani Conformer-L, 17 blocks, $d{=}512$, 8 heads, kernel 31, rel.\ pos.; dropout 0.1 \\
Layer mix & learned $\bm{\alpha}=\operatorname{softmax}(\mathbf{w})$ over all 17 blocks \\
Pooling / embed / classes & ASP (1{,}024-d) $\to$ 256-d $\to$ 23-way \\
Loss & focal $\gamma{=}2$, label smoothing $0.1$ \\
Sampling & $p_l\propto n_l^{1/T}$, $T{=}1.75$; EN floor 16\%, HI floor 11\%; competence curriculum over 17 sub-epochs raising the weight of the 15 Indo-Aryan languages from 0.45 to 1 after a plain-mix first sub-epoch \\
Segments & 0.25--16\,s; 50\% cropped, of which 50\% log-uniform in $[0.25,3]$\,s and the rest uniform in $[3,16]$\,s \\
Augmentation & speed 0.95--1.05 ($p{=}0.5$); MUSAN noise 0--15\,dB ($p{=}0.5$); RIR ($p{=}0.3$); $\mu$-law 8\,kHz codec ($p{=}0.3$); babble of 1--6 talkers at 5--18\,dB ($p{=}0.2$) \\
SpecAugment & 2 frequency masks (width 20 bins), 2 time masks (up to 5\% of frames each) \\
Optimiser & AdamW, $\beta{=}(0.9,0.98)$, weight decay $10^{-3}$ \\
Peak LR / schedule & $3{\times}10^{-4}$; 5\% linear warm-up, cosine decay to $10^{-6}$ \\
Precision / EMA & bf16; EMA decay $0.9999$ \\
Batch / steps / hardware & 144 utterances (6 GPUs $\times$ 24); 425k steps; 6$\times$ RTX A6000, about 70\,h \\
Data & 54{,}643\,h / 31.24M utterances / 23 classes \\
Parameters & 115.6M \\
\bottomrule
\end{tabular}
\caption{Full-scale (Stage~VI) training settings.}
\label{tab:hparams}
\end{table*}

\section{Compared systems: versions and dates}
\label{app:systems}

All open and commercial systems received the same 47{,}730 core clips and the same 17{,}229 all-23 clips (Table~\ref{tab:systems}). Prompted audio LLMs and Azure were given the candidate list (10 languages on core conditions; 23 on the all-23 panel). Sarvam Saaras was queried in auto-detect mode. Open models predicted over their native inventories. \sys always predicted over all 23 classes except in the ``matched'' row of Table~\ref{tab:panel}. All API queries were made between 16 and 21 July 2026.

\begin{table*}[t]
\centering
\small
\setlength{\tabcolsep}{4pt}
\begin{tabular}{@{}>{\raggedright\arraybackslash}p{0.19\textwidth} >{\raggedright\arraybackslash}p{0.28\textwidth} >{\raggedright\arraybackslash}p{0.46\textwidth}@{}}
\toprule
System & Identifier / snapshot & Notes \\
\midrule
Gemini 3.5 Flash & \texttt{google/gemini-3.5-flash} & OpenRouter; card dated 19 May 2026 \\
GPT-audio & \texttt{openai/gpt-audio} & OpenRouter; family snapshot 2025-08-28 \\
GPT-audio-mini & \texttt{openai/gpt-audio-mini} & OpenRouter; snapshots 2025-10-06 / 2025-12-15 \\
Azure Speech LID & Azure AI Speech LID & at most 10 candidates; 11 of 23 locales \\
Sarvam Saaras v3 & \texttt{saaras:v3} & auto-detect; 22 scheduled + \texttt{en-IN} \\
MMS-LID family & \texttt{facebook/mms-lid-*} & 126, 256, 512, 1024, 2048, 4017 \\
NeMo AmberNet & NVIDIA NeMo langid AmberNet & 15 of 23 classes \\
VoxLingua107 ECAPA & SpeechBrain VL107 & 15 of 23 classes \\
Whisper-large-v3 & OpenAI Whisper large-v3 & 15 of 23 classes \\
Voxlect Indic-Whisper & Voxlect Indic-Whisper-v3 & 23 classes; trained on 3--15\,s audio \\
\bottomrule
\end{tabular}
\caption{Systems scored on \panel. API runs used the listed OpenRouter or vendor identifiers. Open models run locally. Commercial behaviour may change between versions; these identifiers are the ones we queried.}
\label{tab:systems}
\end{table*}

\section{Additional results}
\label{app:panel}

\paragraph{Top-1 accuracy.} Mean top-1 accuracy over the 37 core conditions is 0.895 for \sys (0.915 matched); Gemini 3.5 Flash 0.858; Sarvam Saaras v3 0.865; GPT-audio 0.732; Azure Speech LID 0.669; GPT-audio-mini 0.657; MMS-LID-126 0.806; MMS-LID-256 0.789; MMS-LID-512 0.777; MMS-LID-1024 0.772; MMS-LID-2048 0.785; MMS-LID-4017 0.778; NeMo AmberNet 0.771; Voxlect Indic-Whisper 0.726; VoxLingua107 ECAPA 0.745; Whisper-large-v3 0.684. Sarvam's top-1 accuracy exceeds Gemini's although its macro-F1 is lower, which indicates that its errors are concentrated in a few languages.

\paragraph{Further MMS-LID variants.} MMS-LID-512 and MMS-LID-1024 were omitted from Table~\ref{tab:panel} for space. Both cover 21 of the 23 classes. On the 37 core conditions they score 0.808 and 0.801 mean macro-F1 (top-1 0.777 and 0.772). All-23 macro-F1, computed only over the languages each model can emit, is 0.473 and 0.452. Paired differences versus \sys are 10.6 points (95\% CI $[10.1, 11.1]$; W/L $29/1$) and 11.3 points ($[10.9, 11.7]$; $29/1$). They sit between MMS-LID-256 and MMS-LID-2048 on the same ranking; enlarging the MMS-LID inventory beyond 126 languages does not help this 23-way Indic set.

\paragraph{Where Gemini leads.} Apart from the two Hindi-variety probes discussed in \S\ref{sec:errors} and the \panel FLEURS open-set slice, which contains only in-set clips and therefore duplicates FLEURS clean, Gemini 3.5 Flash is significantly ahead of \sys on six core conditions (McNemar, Bonferroni-corrected over 37; Table~\ref{tab:panel_full_a}): the five full-length FLEURS slices and MUCS read speech. Macro-F1 is 0.961 against 0.977 on clean FLEURS, 0.954 against 0.972 with reverberation, 0.951 against 0.966 at 0\,dB, 0.913 against 0.970 over telephony and 0.959 against 0.978 on the confusable subset; on MUCS it is 0.974 against 0.986. These are long, read, high-register clips, and the same sources reverse the ranking once they are cropped to one second.

\paragraph{Calibration and layers.} Figure~\ref{fig:calibration} shows reliability before and after temperature scaling, and Figure~\ref{fig:layers_full} gives the full layer analysis of \S\ref{sec:layers}.

\paragraph{Per-condition results.} Table~\ref{tab:panel_full_a} gives the macro-F1 of \sys and the commercial APIs on all 42 conditions.

\begin{table*}[t]
\centering
\scriptsize
\setlength{\tabcolsep}{3.2pt}
\renewcommand{\arraystretch}{0.92}
\begin{tabular}{@{}l l r r r r r r r r@{}}
\toprule
Domain & Condition & $n$ & \sys & matched & Gemini & Sarvam & GPT-aud. & GPT-mini & Azure \\
\midrule
Clean read & FLEURS & 1{,}500 & .961 & .963 & \textbf{.977} & .864 & .932 & .845 & .773 \\
 & Kathbath & 1{,}500 & \textbf{.987} & .987 & .963 & .864 & .917 & .840 & .795 \\
 & Common Voice & 969 & .921 & .920 & .921 & .804 & .828 & .741 & .729 \\
 & IndicTTS (studio) & 800 & .978 & .979 & .988 & \textbf{.998} & .943 & .923 & .777 \\
 & SPRING-INX & 1{,}200 & .953 & .955 & .865 & \textbf{.956} & .700 & .622 & .686 \\
 & Production mix & 1{,}500 & \textbf{.979} & .981 & .923 & .860 & .842 & .746 & .818 \\
 & MUCS & 750 & .974 & .995 & \textbf{.986} & .975 & .892 & .823 & .888 \\
 & FLEURS open-set$^a$ & 1{,}500 & .961 & .963 & \textbf{.977} & .864 & .932 & .845 & .773 \\
\addlinespace[1pt]
Short & FLEURS 0.5\,s & 1{,}500 & \textbf{.545} & .546 & .231 & .439 & .033 & .000 & .215 \\
 & FLEURS 1\,s & 1{,}500 & \textbf{.764} & .766 & .520 & .667 & .139 & .061 & .456 \\
 & FLEURS 1\,s (VAD) & 1{,}500 & \textbf{.818} & .823 & .589 & .722 & .135 & .068 & .482 \\
 & FLEURS 3\,s & 1{,}500 & \textbf{.956} & .957 & .893 & .847 & .704 & .559 & .806 \\
 & In-domain 1\,s & 1{,}500 & \textbf{.854} & .866 & .611 & .767 & .159 & .090 & .519 \\
 & In-domain 3\,s & 1{,}500 & \textbf{.971} & .975 & .902 & .859 & .735 & .625 & .830 \\
 & In-domain 5\,s & 1{,}411 & \textbf{.984} & .987 & .935 & .863 & .869 & .786 & .836 \\
 & Kathbath 1\,s & 1{,}500 & \textbf{.847} & .855 & .583 & .721 & .141 & .080 & .490 \\
\addlinespace[1pt]
Telephony & FLEURS & 1{,}500 & .913 & .914 & \textbf{.970} & .862 & .905 & .828 & .758 \\
 & Kathbath & 1{,}500 & \textbf{.962} & .977 & .958 & .864 & .901 & .804 & .798 \\
 & IndicVoices & 1{,}500 & \textbf{.921} & .920 & .888 & .847 & .768 & .699 & .726 \\
 & SPRING-INX & 1{,}200 & .938 & .940 & .845 & \textbf{.954} & .674 & .598 & .669 \\
\addlinespace[1pt]
Noise 0\,dB & FLEURS & 1{,}500 & .951 & .954 & \textbf{.966} & .860 & .881 & .790 & .721 \\
 & Kathbath & 1{,}500 & \textbf{.983} & .983 & .943 & .862 & .816 & .722 & .751 \\
 & IndicVoices & 1{,}500 & \textbf{.925} & .927 & .839 & .833 & .709 & .629 & .669 \\
 & SPRING-INX & 1{,}200 & \textbf{.919} & .923 & .788 & .914 & .628 & .531 & .601 \\
\addlinespace[1pt]
Reverb & FLEURS & 1{,}500 & .954 & .958 & \textbf{.972} & .863 & .921 & .829 & .702 \\
 & Kathbath & 1{,}500 & \textbf{.982} & .983 & .964 & .864 & .892 & .802 & .729 \\
 & IndicVoices & 1{,}500 & \textbf{.927} & .929 & .864 & .839 & .755 & .669 & .682 \\
 & SPRING-INX & 1{,}200 & .929 & .926 & .818 & \textbf{.937} & .658 & .575 & .631 \\
\addlinespace[1pt]
Babble & FLEURS & 1{,}500 & \textbf{.905} & .908 & .897 & .808 & .798 & .706 & .657 \\
 & IndicVoices & 1{,}500 & \textbf{.915} & .918 & .783 & .780 & .668 & .592 & .664 \\
\addlinespace[1pt]
Confusable & In-domain & 1{,}350 & \textbf{.978} & .982 & .926 & .850 & .848 & .773 & .842 \\
 & FLEURS & 1{,}350 & .959 & .960 & \textbf{.978} & .849 & .926 & .834 & .765 \\
 & Kathbath & 1{,}350 & \textbf{.987} & .987 & .960 & .849 & .916 & .838 & .817 \\
\addlinespace[1pt]
Hindi varieties & GramVaani (rural tel.) & 150 & .997 & 1.00 & .990 & \textbf{1.00} & .940 & .990 & .305 \\
 & Lahaja (dialectal) & 150 & .790 & .824 & .947 & \textbf{.958} & \textbf{.958} & .929 & .592 \\
 & HiACC (Hinglish) & 150 & .551 & .729 & .800 & .795 & \textbf{.824} & .805 & .723 \\
\addlinespace[1pt]
Hard & Kathbath hard & 1{,}500 & \textbf{.984} & .984 & .954 & .865 & .892 & .787 & .762 \\
\addlinespace[1pt]
All-23 panel & Clean & 3{,}450 & \textbf{.943} & .943 & .831 & .376 & .540 & .420 & .752 \\
 & 1\,s & 3{,}450 & \textbf{.777} & .777 & .367 & .293 & .032 & .018 & .456 \\
 & 3\,s & 3{,}429 & \textbf{.964} & .964 & .779 & .355 & .371 & .265 & .788 \\
 & Telephony & 3{,}450 & \textbf{.894} & .894 & .820 & .359 & .514 & .393 & .737 \\
 & Noise 0\,dB & 3{,}450 & \textbf{.907} & .907 & .753 & .350 & .464 & .331 & .643 \\
\midrule
 & Core-37 mean &  & .914 & .923 & .863 & .847 & .735 & .659 & .687 \\
 & All-23 mean &  & .897 & -- & .710 & .346 & .384 & .285 & -- \\
\bottomrule
\end{tabular}
\caption{Per-condition macro-F1 on \panel for \sys and the commercial APIs (identical clips). \emph{matched}: \sys restricted to the candidate set given to the prompted APIs. Bold marks the best of all 16 systems scored, open models included. $^a$The panel's FLEURS open-set slice contains only in-set clips and duplicates FLEURS clean. Azure supports 11 of the 23 classes and is omitted from the all-23 rows.}
\label{tab:panel_full_a}
\end{table*}

\end{document}